\pdfoutput=1

\documentclass[11pt]{article}

\usepackage[preprint]{acl}

\usepackage{times}
\usepackage{latexsym}

\usepackage[T1]{fontenc}
\usepackage{enumitem}
\usepackage{subcaption}
\usepackage[utf8]{inputenc}
\usepackage{tabularx}
\usepackage{adjustbox}
\usepackage{microtype}

\usepackage{inconsolata}
\usepackage{booktabs}
\usepackage{graphicx}
\usepackage{subcaption}
\usepackage{tikz}
\usepackage{multicol}
\usetikzlibrary{trees}
\usetikzlibrary{arrows.meta,positioning,fit,shadows.blur}
\usepackage{hyperref}

\title{The Role of Dataset Linguistic Structure in the Cultural\\Awareness of Large Language Models}

\author{
  \textbf{Reem I. Masoud\textsuperscript{1,5}\thanks{Corresponding authors: 
  \href{mailto:reem.masoud.22@ucl.ac.uk}{reem.masoud.22@ucl.ac.uk}, 
  \href{mailto:c.feng@qub.ac.uk}{c.feng@qub.ac.uk}}},
  \textbf{Chen Feng\textsuperscript{2,1}\footnotemark[1]},
  \textbf{Shunta Asano\textsuperscript{3,1}},\\
  \textbf{Saied Alshahrani\textsuperscript{4}},
  \textbf{Philip Colin Treleaven\textsuperscript{1}},
  \textbf{Miguel R. D. Rodrigues\textsuperscript{1,6}}
\\
\\
  \textsuperscript{1}University College London\\
  \textsuperscript{2}Queen's University Belfast,
  \textsuperscript{3}The University of Tokyo,\\
  \textsuperscript{4}University of Bisha,
  \textsuperscript{5}King Abdulaziz University,\\
  \textsuperscript{6}AI Centre, University College London
}

\begin{document}
\maketitle
\begin{abstract}
The global deployment of large language models (LLMs) has raised concerns about cultural misalignment, yet the linguistic properties of fine-tuning datasets used for cultural adaptation remain poorly understood. We adopt a dataset-centric view of cultural alignment and investigate which properties of post-training data are associated with cultural performance, whether they can guide data selection before fine-tuning, and how their effects vary across languages and model families. We compute lightweight linguistic, semantic, and structural metrics for Arabic, Chinese, and Japanese datasets and apply principal component analysis (PCA) separately within each language.
The resulting components form broadly interpretable axes: PC1 is generally dominated by semantic structure, PC2 captures diversity and lexical variation, and PC3 reflects more language-specific organization. We fine-tune LLaMA, Mistral, and DeepSeek models and evaluate them on benchmarks of cultural knowledge, values, and norms. Although the PCA-derived dataset descriptors are associated with downstream performance, the strongest relationships vary across models, benchmarks, and languages, indicating that no single component serves as a universal predictor. Controlled, size-matched subset interventions further show that PCA-guided selection can improve cultural performance when the relevant component and direction are validated against random sampling. PC3 provides the strongest signal for Arabic, while High-PC1 is most effective for Japanese, particularly for LLaMA. Chinese results are weaker and more model-specific and remain exploratory because of smaller subset sizes. Overall, our findings show that lightweight dataset descriptors can support pre-training data diagnostics, but effective cultural adaptation requires language- and architecture-aware selection rather than a universal linguistic criterion. 

\end{abstract}

\section{Introduction}

LLMs have shown remarkable progress across diverse natural language processing (NLP) tasks. However, their performance often falls short in cross-cultural settings, where linguistic variation, cultural norms, and local knowledge shape how users interpret and engage with model outputs~\citep{prabhakaran2022cultural}. Cultural alignment, which ensures that LLMs understand and reflect the values, norms, and nuances of the user groups interacting with them~\cite{masoud2023cultural}, is therefore essential for building inclusive, globally applicable AI systems. While recent work highlights the importance of dataset quality for model performance~\citep{zhou2023lima, alshahrani2023performance}, cultural datasets remain understudied from a linguistic and structural perspective.

Dominant approaches to cultural alignment focus on value congruence, relying on survey-derived benchmarks (e.g., World Values Survey) or synthetic QA pairs, with limited attention to the linguistic structure of the underlying training data. As a result, these approaches often overlook how cultural nuances are encoded in the structural, semantic, and stylistic properties of the data, and whether such properties can be assessed independently of model training.

Cultural datasets also remain under-represented for many non-Western languages~\citep{pawar2025survey}, and it remains unclear how their linguistic properties influence model behavior. Prior work further suggests that different model architectures may respond differently to the same training data~\citep{yauney2023data, zhang2025unveiling}.
We therefore ask whether cultural fine-tuning datasets exhibit consistent, measurable patterns in semantic content, lexical diversity, and stylistic variability that can be quantified prior to training, and whether such properties are predictive of, and actionable for, downstream cultural behavior across different model families. To answer these questions, we (i) characterize multilingual datasets using lightweight linguistic metrics and PCA, (ii) test associations between PCA dimensions and cultural benchmark performance across models, and (iii) probe actionability through controlled high/low/random subset fine-tuning. Our contributions are threefold:

\begin{enumerate}
\item We present a \emph{dataset-centric methodology} that quantifies linguistic, semantic, and structural properties of cultural datasets, reduces them via PCA, and links these components to downstream cultural alignment performance.
\item To our knowledge, we conduct the \emph{first} cross-lingual empirical study examining Arabic, Chinese, and Japanese datasets across three LLM families, revealing that correlations between dataset properties and cultural performance vary substantially by language and model architecture.
\item We evaluate the predictive and practical utility of PCA-derived linguistic dimensions for dataset assessment, including subset-based interventions that test whether these signals remain informative under controlled data selection.
\end{enumerate}

Overall, our results indicate that dataset linguistic properties provide useful signals for cultural alignment, but their effects vary substantially across languages and model architectures. This motivates language- and model-aware, dataset-centric strategies for selecting post-training data for culturally aligned LLMs.

\section{Related Work}
Recent work has introduced datasets and benchmarks for evaluating cultural behaviour in LLM, particularly value-based resources derived from surveys comparing model outputs to human responses~\cite{alkhamissi2024investigating}. 
These approaches provide useful evaluation frameworks, but they primarily assess behavioural outcomes rather than examining the linguistic characteristics of the datasets that may influence those outcomes.
A complementary line of research studies cultural bias in pre-training data. 
Naous et al.~\cite{naous2023having} show that widely used corpora such as Wikipedia exhibit strong Western-centric bias, and broader surveys document representational imbalance across languages and cultures~\cite{pawar2025survey}. 
While these works identify important dataset-level biases, they primarily focus on representation and coverage rather than measuring linguistic structure or testing how such properties relate to downstream cultural alignment.
Other work proposes methods for generating or curating culturally diverse datasets. 
For example, CulturePark~\cite{li2024culturepark} generates synthetic culturally diverse dialogues and studies downstream effects. 
However, these approaches focus on dataset construction and augmentation rather than identifying which dataset-level properties are associated with culturally appropriate model behaviour.
Beyond cultural applications, dataset-centric studies such as DSAP~\cite{dominguez2025dsap} analyze demographic similarity across corpora. 
Related work on dataset quality and diversity examines how curation, filtering, and diversity-oriented selection strategies influence downstream performance~\cite{chen2024diversity, zhou2023lima, pang2024improving}. 
These studies demonstrate that dataset properties matter for downstream behaviour, but they do not directly connect measurable linguistic and structural properties of post-training data to cultural performance on culture-specific benchmarks.

In contrast, out work investigates which measurable linguistic properties of post-training datasets, including semantic and structural characteristics, are associated with culturally appropriate behaviour.
We introduce a dataset-centric methodology that computes lightweight dataset metrics across multilingual fine-tuning datasets and reduces them via PCA.
We then relate these properties to cultural performance across models and languages, revealing systematic patterns and highlighting the potential for evaluating dataset utility \textit{prior} to model training.
\begin{figure}[ht]
    \centering
    \includegraphics[width=1.05\linewidth]{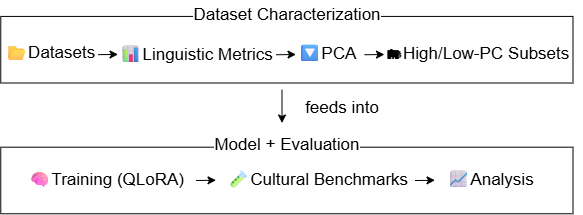}
    \caption{Overview of the proposed pipeline.} \label{fig:methodology}
\end{figure}

\section{Methodology}
\label{sec:methodology}

Our methodology follows the two-stage pipeline shown in Figure~\ref{fig:methodology}: 
\emph{Dataset Characterization} and \emph{Model Evaluation}. 

\paragraph{Dataset Characterization.}
For each language, we characterize every source fine-tuning dataset using complementary linguistic, semantic, and structural metrics. \emph{Surface diversity} is measured using Distinct-1, Distinct-2~\citep{li2016diversity}, and Self-BLEU~\citep{papineni-etal-2002-bleu}, which capture lexical and phrasal variation as well as repetition across examples. \emph{Lexical richness} is measured using type-token ration (TTR)~\citep{johnson1944studies}, moving TTR (MATTR)\citep{covington2010cutting}, measure of textual lexical diversity (MTLD)~\citep{mccarthy2010mtld}, and hypergeometric distribution diversity (HDD)~\citep{wu1993accurate,mccarthy2010mtld}, which quantify the breadth and stability of vocabulary use. \emph{Semantic similarity} is estimated using cosine similarity over both TF--IDF representations~\citep{salton1975vector,salton1988term} and transformer-based sentence embeddings~\citep{reimers2019sentence}, allowing us to capture lexical overlap and broader similarity in meaning. \emph{Structural cohesion} is evaluated by clustering sentence embeddings using K-means~\citep{mcqueen1967some} and computing the silhouette score~\citep{rousseeuw1987silhouettes}, indicating how clearly the dataset is organized into internally coherent groups. 
Metrics are aggregated at the dataset level so that each dataset is
represented by a single linguistic profile.
To control for differences in dataset size, we compute all features from a random sample of 1,000 examples per dataset.  Pilot checks using larger samples and multiple
random seeds produced only minimal variation in the resulting feature
values. Because metric distributions may differ across languages due to morphology, script, and tokenization, we construct a separate dataset--feature matrix for each language and standardize each feature across the datasets within that language.

We then apply PCA independently to each standardized language-specific matrix, where rows represent datasets and columns represent metrics. PCA transforms the correlated features into orthogonal components that summarize the main dimensions of variation among datasets. We retain the first three principal components and assign each dataset a score on each component. These scores are used both to characterize complete datasets and to examine whether dataset composition is associated with downstream cultural performance.

To construct controlled fine-tuning subsets, we select, for each language and principal component, the five datasets with the largest signed component scores. Since PCA is fitted at the dataset level, its scores cannot be assigned directly to individual examples. We therefore compute a PCA-inspired proxy score using three features that can be measured per example: TTR, Distinct-1, and transformer-based semantic similarity. For each example \(i\) in language \(\ell\) and principal component \(k\),
we compute the proxy ranking score
\begin{equation}
a_{\ell,i,k}
=
\sum_{q\in\mathcal{Q}}
x_{\ell,i,q}v_{\ell,k,q},
\end{equation}
where
\(\mathcal{Q}=\{\mathrm{TTR},\mathrm{D1},\mathrm{ST}\}\),
\(x_{\ell,i,q}\) is the example-level value of feature \(q\), and
\(v_{\ell,k,q}\) is the loading of feature \(q\) on component \(k\) in the
language-specific PCA solution. Here, \(\mathrm{D1}\) denotes Distinct-1
and \(\mathrm{ST}\) denotes sentence-transformer semantic similarity.
The score \(a_{\ell,i,k}\) is used only to rank examples within the same
source dataset and is not interpreted as a true example-level PCA
coordinate. For each selected dataset, the \textsc{High-PC} subset contains examples from the upper tail of the ranking, the \textsc{Low-PC} subset contains examples from the lower tail, and a size-matched \textsc{Random} subset serves as the baseline. Subset sizes are adjusted to preserve non-overlapping tails, and random examples are drawn from unused examples whenever possible using a fixed seed.

\paragraph{Model Evaluation.}
For each model and language, we fine-tune a separate copy of the same pre-trained checkpoint on every complete source dataset. For the controlled subset intervention, we additionally fine-tune separate copies on the \textsc{High-PC}, \textsc{Low-PC}, and size-matched \textsc{Random}
subsets constructed from the selected source datasets identified in the previous section. 
Each run starts independently from the same pre-trained checkpoint; no fine-tuned model is used to initialize a subsequent run. Preprocessing, prompt formatting, QLoRA configuration, optimisation settings, and training duration are held constant across comparable runs so that performance differences can be attributed to the fine-tuning data used. Each fine-tuned model is evaluated on language-specific benchmarks covering cultural knowledge, cultural values, and cultural norms. We use the standard task-specific metric for each benchmark and direction-align scores where necessary so that larger values consistently indicate better cultural performance.

We analyse the results at two levels. First, for models fine-tuned on complete datasets, we compute Pearson correlations between each dataset's PCA score and its downstream benchmark performance. This analysis tests whether the dataset-level dimensions identified before training are associated with subsequent cultural performance, but is interpreted as descriptive rather than causal. Second, we conduct a controlled subset intervention by comparing the performance of models fine-tuned on \textsc{High-PC} and \textsc{Low-PC} subsets with the performance of the corresponding \textsc{Random} baseline. We summarise these comparisons using benchmark-level wins, ties, and losses, together with the mean performance difference normalised by each benchmark's observed score range. This intervention tests whether the PCA-derived ranking identifies examples that affect cultural performance beyond the effect of subset size alone. 

\section{Experiments}
\label{sec:experiments}

\subsection{Experimental Setup}

\paragraph{Languages, Datasets, and Evaluation.}
We evaluate our approach in Arabic, Japanese, and Chinese using 9--13
post-training datasets per language. These languages provide a multilingual
testbed spanning different scripts, linguistic structures, and cultural
contexts, while offering sufficient publicly available fine-tuning datasets
and cultural benchmarks for systematic comparison. The datasets cover
question answering, examinations, social media, news, instruction tuning,
web text, lexical resources, and curated cultural data. All datasets are
converted into supervised input--output pairs using a common
prompt--response format. Cultural performance is evaluated along three
dimensions: \emph{Cultural Knowledge}, \emph{Cultural Values}, and
\emph{Cultural Norms}. We use the standard task metric for each benchmark
and direction-align scores so that larger values consistently indicate
better performance.
Table~\ref{tab:data_benchmarks_summary} summarizes the datasets and
benchmarks used for each language; detailed dataset statistics and the
complete benchmark breakdown are provided in
Appendices~\ref{appendix:dataset_stats} and~\ref{appendix:benchmarks},
respectively.

\begin{table*}[t]
\centering
\footnotesize
\setlength{\tabcolsep}{4pt}
\begin{tabularx}{\textwidth}{
    p{1.1cm}
    >{\raggedright\arraybackslash}X
    >{\raggedright\arraybackslash}X}
\toprule
\textbf{Lang.} &
\textbf{Post-training datasets} &
\textbf{Cultural evaluation benchmarks} \\
\midrule

AR &
MultiNativQA~\citep{hasan2024nativqa},
ArabicMMLU~\citep{koto2024arabicmmlu},
Aya$^{*}$~\citep{singh2024aya},
CIDAR~\citep{alyafeai2024cidar},
Open-ArabicaQA~\citep{abdallah2024arabicaqa},
Ultimate Arabic News~\citep{ultimatenews},
DAWQAS~\citep{ismail2018dawqas},
Al Jazeera News~\citep{arbml_aljazeera_news_articles},
and Arsen-20~\citep{arbml_arsen20}
&
\textit{Knowledge:}
CulturalBench-Easy/Hard~\citep{chiu2024culturalbench},
CultureAtlas~\citep{fung2024massively},
ArabCulture~\citep{sadallah2025arabculture},
and EXAMs-AR~\citep{sengupta2023jais};
\textit{Values:}
VSM13~\citep{hofstede2013vsm},
WorldValuesBench~\citep{zhao2024worldvaluesbench},
and ACVA-Arabic~\citep{huang2023acegpt};
\textit{Norms:}
CIDAR-MCQ/EVAL~\citep{alyafeai2024cidar},
AraDiCE~\citep{mousi2024aradice},
and LLM-Globe-Open/Closed~\citep{karinshak2024llm}
\\

\midrule

JP &
Ichikara~\citep{sekine2024ichikara},
Databricks-Dolly-JA~\citep{kunishou_dolly15k_ja},
JaQuAD~\citep{so2022jaquad},
Japanese WordNet~\citep{bond-kuribayashi-2023-japanese},
JCommonsense~\citep{takeshita2023jcommonsense},
AIO~\citep{aio_dataset},
Easy-Japanese~\citep{katsuta2018simplification},
Japanese Wikipedia~\citep{wikipedia_ja_20230720,wikimedia_dumps},
Wikinews~\citep{Hirano2023-llmj,wikinews_japanese},
and Aozora Bunko~\citep{aozora_txt}
&
\textit{Knowledge:}
CulturalBench-Easy/Hard~\citep{chiu2024culturalbench};
\textit{Values:}
VSM13~\citep{hofstede2013vsm}
and WorldValuesBench~\citep{zhao2024worldvaluesbench};
\textit{Norms:}
J-Ethics~\citep{takeshita2025jethicsjapaneseethicsunderstanding}
and LLM-Globe-Open/Closed~\citep{karinshak2024llm}
\\

\midrule

ZH &
COIG-PC, Chinese\_Traditional, Douban, Xiaohongshu,
Human\_Value, RuozhiBa, Exam, SegmentFault, Zhihu, LogiQA,
WikiHow, Wiki, and Finance~\citep{bai2024coig}
&
\textit{Knowledge:}
CulturalBench-Easy/Hard~\citep{chiu2024culturalbench},
CultureAtlas~\citep{fung2024massively},
and CEval-Exams~\citep{huang2023ceval};
\textit{Values:}
VSM13~\citep{hofstede2013vsm}
and WorldValuesBench~\citep{zhao2024worldvaluesbench};
\textit{Norms:}
CMoralEval~\citep{yu2024cmoralevalmoralevaluationbenchmark}
and LLM-Globe-Open/Closed~\citep{karinshak2024llm}
\\

\bottomrule
\end{tabularx}
\caption{Post-training datasets and cultural evaluation benchmarks used
for each language. Detailed dataset sizes, domains, and source types are
provided in the supplementary material. $^{*}$Arabic subset obtained through
language filtering.\centering}
\label{tab:data_benchmarks_summary}
\end{table*}

\paragraph{Models and Training.}
We fine-tune Llama-3.2-3B-Instruct, Mistral-7B-Instruct-v0.3, and
DeepSeek-R1-Distill-Qwen-7B. Using three model families allows us to test
whether sensitivity to dataset composition transfers across architectures
and pre-training histories. Datasets containing more than 30,000 eligible
examples are capped at 30,000 after dataset-specific filtering and seeded
shuffling; smaller datasets are used in full. Each dataset is split into
80\% training, 10\% validation, and 10\% test partitions using a fixed
seed. All models are fine-tuned for three epochs using 4-bit QLoRA with
identical optimisation settings. Table~\ref{tab:setup} summarizes the
training configuration. For embedding-based features, we use
\texttt{bert-large-arabertv02}, \texttt{sentence-bert-base-ja-mean-tokens},
and \texttt{chinese-macbert-base} for Arabic, Japanese, and Chinese,
respectively; K-means uses the \(k\in\{2,\ldots,10\}\) that maximizes the
silhouette score.

\begin{table}[htbp]
\centering
\small

\setlength{\tabcolsep}{4pt}
\begin{tabular}{ll}
\toprule
\textbf{Setting} & \textbf{Value} \\
\midrule
Fine-tuning & QLoRA, 4-bit quantization \\
Batch / accumulation & 8 / 8 \\
Learning rate & $2\times10^{-5}$ \\
Optimizer & AdamW \\
Warmup ratio & 0.03 \\
Maximum length & 2,048 tokens \\
Epochs & 3 \\
LoRA $(r,\alpha)$ & $(64,16)$ \\
LoRA dropout & 0.05 \\
Hardware & 2$\times$A100 80GB \\
\bottomrule
\end{tabular}
\caption{Training configuration used for all fine-tuning experiments.\centering}
\label{tab:setup}
\end{table}

\subsection{Linguistic Structure of Fine-Tuning Datasets}
\label{sec:pca_results}
The first three language-specific principal components explain approximately 84\% of the variance in Arabic, 96\% in Japanese, and 88\% in Chinese (Table~\ref{tab:pca_variance}), supporting their use as compact descriptors of dataset composition.

\begin{table}[t]
\centering
\small
\setlength{\tabcolsep}{5pt}
\begin{tabular}{lccc}
\toprule
\textbf{Language} & \textbf{PC1} & \textbf{PC2} & \textbf{PC3} \\
\midrule
Arabic   & 0.52 & 0.17 & 0.15 \\
Japanese & 0.52 & 0.34 & 0.10 \\
Chinese  & 0.42 & 0.30 & 0.16 \\
\bottomrule
\end{tabular}
\caption{Explained variance ratios of the first three principal components.\centering}
\label{tab:pca_variance}
\end{table}

\begin{figure*}[t]
    \centering

    \begin{minipage}[t]{0.32\textwidth}
        \centering
        \includegraphics[width=\linewidth]{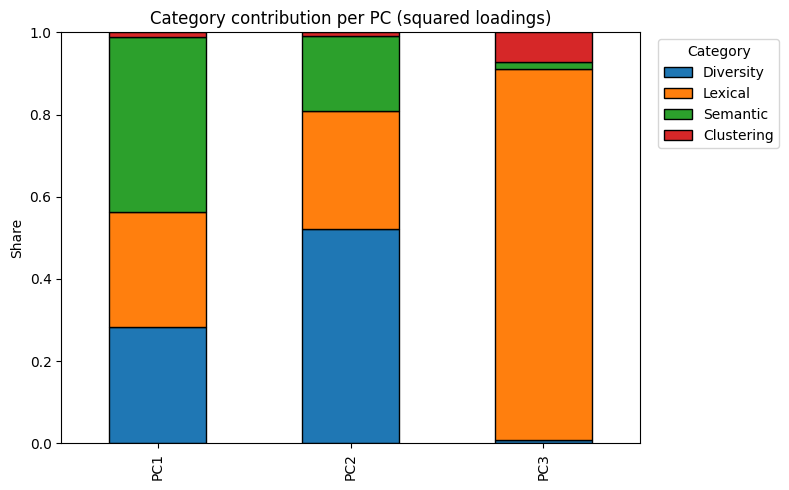}

        \small (a) Arabic
    \end{minipage}
    \hfill
    \begin{minipage}[t]{0.32\textwidth}
        \centering
        \includegraphics[width=\linewidth]{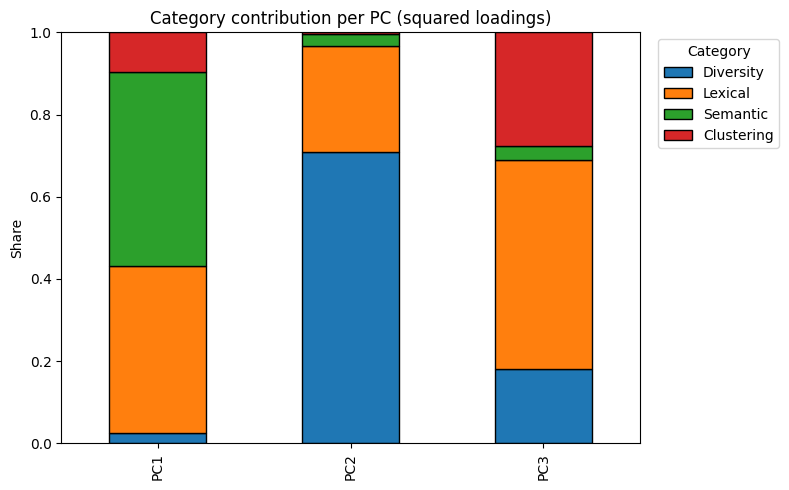}

        \small (b) Japanese
    \end{minipage}
    \hfill
    \begin{minipage}[t]{0.32\textwidth}
        \centering
        \includegraphics[width=\linewidth]{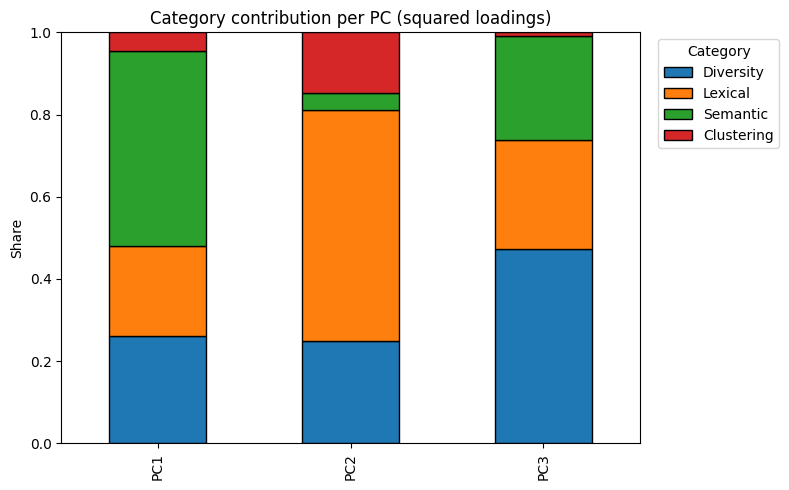}

        \small (c) Chinese
    \end{minipage}

    \caption{Category-level contributions of diversity, lexical,
    semantic, and clustering metrics to PC1--PC3 for
    (a) Arabic, (b) Japanese, and (c) Chinese.}
    \label{fig:cat_all}
\end{figure*}

\paragraph{Component interpretation.}
Figure~\ref{fig:cat_all} shows that the components emphasize coherent but
language-dependent combinations of dataset properties. PC1 is
semantic-dominant across all three languages, although lexical and
diversity features also contribute substantially. We therefore interpret
PC1 as a mixed dimension primarily reflecting meaning-level consistency
and similarity to culture-related semantic content. PC2 is driven mainly
by diversity and lexical features: diversity dominates in Arabic and
Japanese, whereas lexical richness contributes more strongly in Chinese.
It therefore captures variation in wording, vocabulary use, and repeated
surface forms. PC3 is less consistent across languages, reflecting lexical
richness in Arabic, a lexical--clustering contrast in Japanese, and a
secondary diversity-oriented dimension in Chinese. Consequently, PC3 must be
interpreted separately for each language.

\begin{figure}[htbp]
    \centering

    \includegraphics[width=\columnwidth]
    {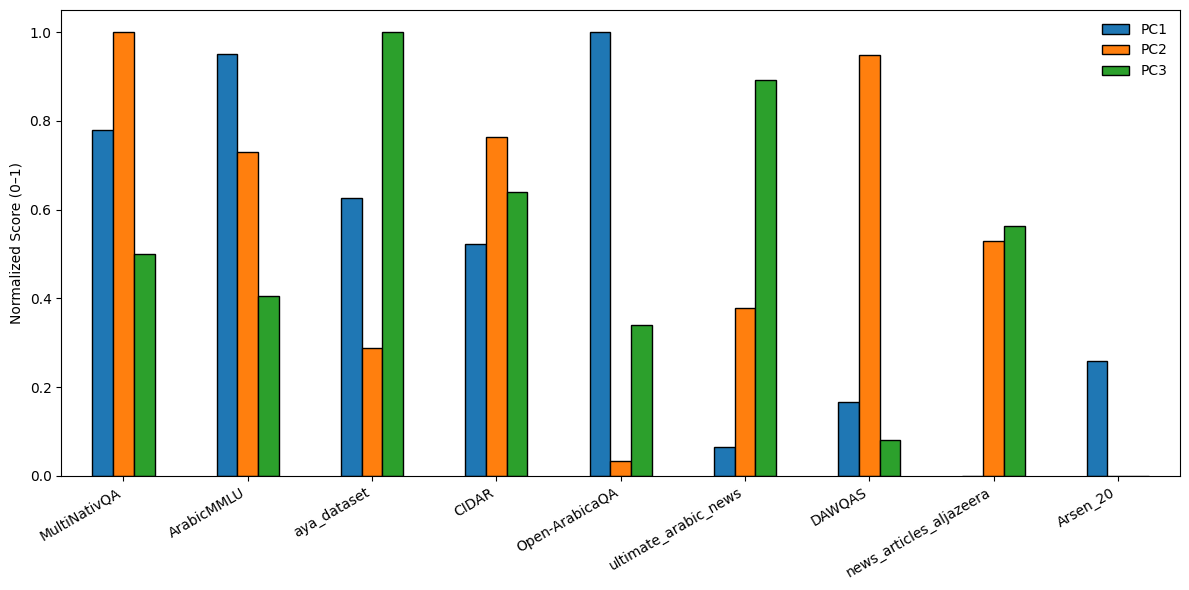}

    \small (a) Arabic

    \vspace{0.6em}

    \includegraphics[width=\columnwidth]
    {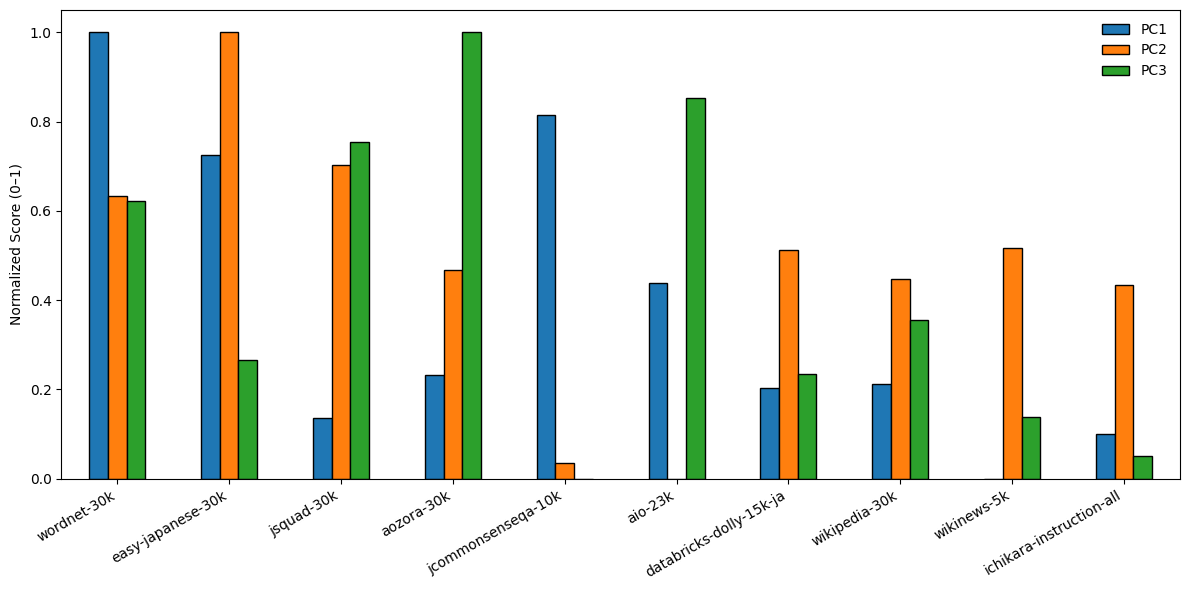}

    \small (b) Japanese

    \vspace{0.6em}

    \includegraphics[width=\columnwidth]
    {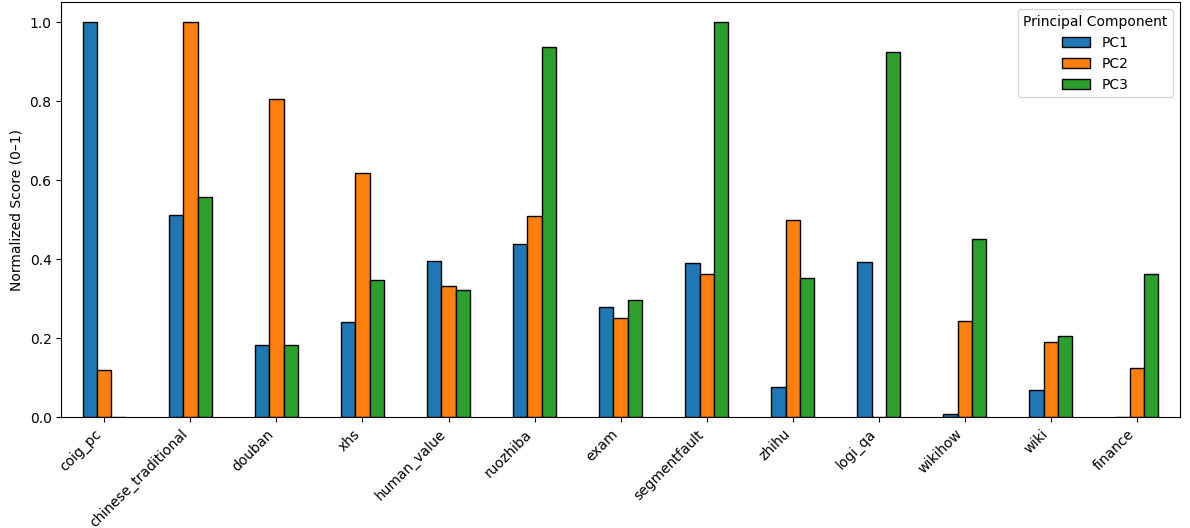}

    \small (c) Chinese

    \caption{Normalized PC1--PC3 scores of the source fine-tuning
    datasets for (a) Arabic, (b) Japanese, and (c) Chinese. Scores show
    the relative position of each dataset in its language-specific PCA
    space and are not directly comparable across languages.}
    \label{fig:scores_all}
\end{figure}

\paragraph{Dataset interpretation.}
The dataset scores in Figure~\ref{fig:scores_all} make the component contrasts more concrete. Higher PC1 scores are generally associated with QA, lexical, and reasoning-oriented resources, such as MultiNativQA,
Japanese WordNet, and COIG-PC, whose examples are organized around targeted questions, definitions, or semantic relations. Lower-PC1 datasets, such as
Ultimate Arabic News, Wikinews, and WikiHow, use more expository or procedural formats. PC2 separates datasets with more varied wording and communicative styles, such as DAWQAS, JaQuAD, and Douban, from more
regularized or constrained resources, such as Arsen-20, JCommonsense, and LogiQA. PC3 is more language dependent: high-scoring examples include Aya
and CIDAR in Arabic, Aozora and WordNet in Japanese, and SegmentFault and RuozhiBa in Chinese, reflecting different combinations of vocabulary use, internal organization, and expression style. These interpretations remain descriptive and do not imply that either end of a component is preferable
for cultural alignment.

\subsection{Dataset Structure and Cultural Performance}
\label{sec:correlation_results}

For each source dataset, we independently fine-tune LLaMA, Mistral, and
DeepSeek and compute Pearson correlations, with 95\% confidence intervals,
between its PC1--PC3 scores and cultural benchmark performance. Given the 9--13 datasets available per language, these correlations and
their confidence intervals are interpreted descriptively.
Selected strong associations are shown in Table~\ref{tab:selected_correlations}; complete coefficient tables and confidence intervals are provided in Appendix~\ref{appendix:additional_results}, with the corresponding heatmaps in Appendix~\ref{appendix:correlation_heatmaps}.

The results show that linguistic structure is associated with downstream
performance, but the relevant dimension depends on the model and task.
For example, Arabic CultureAtlas is most strongly associated with PC1 for
LLaMA, whereas EXAMs and ACVA are associated with PC2 for Mistral and
DeepSeek. Japanese VSM13 shifts from PC2 for LLaMA to PC1 for DeepSeek,
while CulturalBench-Easy is negatively associated with PC1 across model
families. Chinese results are similarly distributed across PC1--PC3.
Thus, no component or direction provides a consistent selection rule across
languages, benchmarks, and architectures, motivating the controlled subset
interventions in the following section.

\begin{table}[t]
\centering
\small
\setlength{\tabcolsep}{3pt}

\begin{adjustbox}{max width=\linewidth}
\begin{tabular}{llllrr}
\toprule
\textbf{Lang.} & \textbf{Model} & \textbf{Benchmark} &
\textbf{PC} & \(\mathbf{r}\) & \textbf{95\% CI} \\
\midrule
AR & LLaMA    & CultureAtlas       & PC1 &  0.85 & [0.75, 0.97] \\
AR & Mistral  & EXAMs              & PC2 &  0.82 & [0.49, 0.96] \\
AR & DeepSeek & ACVA               & PC2 &  0.78 & [0.41, 0.97] \\
\midrule
JP & LLaMA    & VSM13              & PC2 &  0.73 & [0.17, 0.95] \\
JP & Mistral  & CulturalBench-Easy & PC1 & -0.68 & [-0.94, -0.04] \\
JP & DeepSeek & LLM-Globe-Closed   & PC2 &  0.54 & [0.16, 0.92] \\
\midrule
ZH & LLaMA    & CMoralEval         & PC3 &  0.62 & [0.12, 0.86] \\
ZH & Mistral  & LLM-Globe-Closed   & PC1 &  0.77 & [0.28, 0.93] \\
ZH & DeepSeek & WorldValuesBench   & PC1 &  0.69 & [0.11, 0.93] \\
\bottomrule
\end{tabular}
\end{adjustbox}

\caption{Selected strong correlations between dataset component scores and
cultural performance. Values are Pearson's \(r\) with 95\% confidence
intervals; complete results are provided in the supplementary material.\centering}
\label{tab:selected_correlations}

\end{table}

\subsection{Controlled Validation of PCA-Guided Data Selection}
\label{sec:subset_results}

Correlation alone does not establish whether PCA-derived dataset
dimensions can guide fine-tuning-data selection. We therefore conduct
controlled subset interventions for Arabic, Japanese, and Chinese. For
each language and principal component, we select the five datasets with
the largest signed dataset-level component scores. Within each selected
dataset, examples are ranked using the PCA-inspired proxy score defined
in the Methodology, and size-matched \textsc{High-PC},
\textsc{Low-PC}, and \textsc{Random} subsets are constructed. Each subset
independently fine-tunes LLaMA, Mistral, and DeepSeek under the same
training configuration. Most Arabic and Japanese subsets contain
approximately 2,000 examples; smaller datasets use reduced but
condition-matched subsets. Because the Chinese datasets yield
substantially smaller subsets, the Chinese intervention is interpreted as
an exploratory extension. Detailed
dataset-level results for all controlled interventions are provided in
Appendix~\ref{appendix:subset_results}.

We evaluate each intervention in two complementary ways. First, each
\textsc{High-PC} and \textsc{Low-PC} model is compared with its
corresponding \textsc{Random} baseline on every benchmark. A
source-dataset intervention is classified as an improvement when wins
outnumber losses, a decline when losses outnumber wins, and a tie when
the counts are equal. Second, we compute the mean normalized performance
margin relative to Random, which captures the scale-adjusted magnitude of
the intervention effect. Directional outcomes measure how recurrently an
intervention helps across benchmarks, whereas normalized margins measure
the average size of the resulting gains or declines.

\begin{table}[htbp]
\centering
\small
\setlength{\tabcolsep}{5pt}
\begin{tabular}{lcccccc}
\toprule
& \multicolumn{2}{c}{\textbf{PC1}} &
\multicolumn{2}{c}{\textbf{PC2}} &
\multicolumn{2}{c}{\textbf{PC3}} \\
\cmidrule(lr){2-3}
\cmidrule(lr){4-5}
\cmidrule(lr){6-7}
\textbf{Lang.} &
\textbf{High} & \textbf{Low} &
\textbf{High} & \textbf{Low} &
\textbf{High} & \textbf{Low} \\
\midrule
AR
& 7/2/6 & 6/4/5
& 2/6/7 & 3/5/7
& 9/4/2 & 8/4/3 \\

JP
& 8/6/1 & 2/10/3
& 4/7/4 & 4/8/3
& 4/10/1 & 4/10/1 \\

ZH
& 4/9/2 & 3/10/2
& 4/6/5 & 7/3/5
& 2/10/3 & 4/9/2 \\
\bottomrule
\end{tabular}
\caption{Aggregate directional outcomes of PCA-guided subset selection.
For each language, principal component, and subset type
(\textsc{High-PC} or \textsc{Low-PC}), entries report
Improvement/Tie/Decline counts across five source datasets and three model
families, yielding \(5\times3=15\) interventions per subset type.\centering}
\label{tab:subset_directional}
\end{table}

\begin{table*}[htbp]
\centering
\small
\setlength{\tabcolsep}{3pt}
\begin{tabular}{ll l r l r}
\toprule
\textbf{Lang.} & \textbf{Model} &
\multicolumn{2}{c}{\textbf{Largest improvement}} &
\multicolumn{2}{c}{\textbf{Largest decline}} \\
\cmidrule(lr){3-4}\cmidrule(lr){5-6}
& & \textbf{Dataset (selection)} & \textbf{Margin} &
\textbf{Dataset (selection)} & \textbf{Margin} \\
\midrule
AR & LLaMA &
Al Jazeera News (PC3-L) & 0.173 &
CIDAR (PC3-L) & -0.112 \\

AR & Mistral &
CIDAR (PC3-H) & 0.131 &
Aya (PC3-H) & -0.188 \\

AR & DeepSeek &
CIDAR (PC3-L) & 0.117 &
Open-ArabicaQA (PC1-H) & -0.302 \\
\midrule

JP & LLaMA &
WordNet (PC3-L) & 0.159 &
JaQuAD (PC3-L) & -0.124 \\

JP & Mistral &
Wikipedia (PC3-H) & 0.156 &
Databricks-Dolly-JA (PC2-H) & -0.056 \\

JP & DeepSeek &
Databricks-Dolly-JA (PC2-L) & 0.298 &
WordNet (PC1-H) & -0.153 \\
\midrule

ZH & LLaMA &
COIG-PC (PC1-L) & 0.244 &
Zhihu (PC2-H) & -0.116 \\

ZH & Mistral &
LogiQA (PC3-L) & 0.125 &
COIG-PC (PC1-L) & -0.109 \\

ZH & DeepSeek &
Zhihu (PC2-L) & 0.248 &
COIG-PC (PC1-H) & -0.235 \\
\bottomrule
\end{tabular}
\caption{Largest positive and negative dataset-level normalized margins
relative to size-matched Random subsets. H and L denote
\textsc{High-PC} and \textsc{Low-PC} selection, respectively. 
Complete dataset-level normalized margins are provided in Appendix~\ref{appendix:magnitude_results}.\centering}
\label{tab:margin_extremes}
\end{table*}

\paragraph{Directional consistency.}
Table~\ref{tab:subset_directional} summarizes the dataset-level
Improvement/Tie/Decline classifications after pooling the five selected
source datasets across the three model families. These outcomes describe
the recurrence of improvement relative to size-matched Random subsets;
they do not represent statistical significance or the magnitude of the
underlying score differences.

For Arabic, PC3 provides the clearest positive selection pattern.
High-PC3 yields nine improvements, four ties, and two declines, while
Low-PC3 yields eight improvements, four ties, and three declines. Both
directions therefore outperform the more balanced PC1 outcomes and the
predominantly neutral or negative PC2 outcomes. The preferred PC3 tail
nevertheless remains architecture dependent: LLaMA and Mistral respond
more favourably to High-PC3, whereas DeepSeek responds most favourably to
Low-PC3. Arabic PC3 is therefore the strongest overall selection
dimension, but not a universal directional rule across models.

Japanese shows a different pattern. High-PC1 produces eight improvements,
six ties, and only one decline, compared with two improvements, ten ties,
and three declines for Low-PC1. This advantage is driven most strongly by
LLaMA, for which High-PC1 improves all five source-dataset interventions.
PC2 is mixed under both directions, while PC3 is dominated by ties:
High-PC3 and Low-PC3 each produce four improvements, ten ties, and one
decline. Thus, Japanese High-PC1 provides the clearest within-dataset
selection signal, whereas PC3 appears to capture structure that is already
largely retained by Random subsets from the selected source datasets.

The Chinese results are weaker and are treated as exploratory. PC1 is
mostly tie dominated under both tails. PC2 produces the largest number of
Chinese improvements under Low-PC2, with seven improvements, three ties,
and five declines, compared with four improvements, six ties, and five
declines under High-PC2. However, this aggregate advantage is not
consistent across architectures: DeepSeek and LLaMA favour Low-PC2,
whereas Mistral produces four declines under the same direction. PC3 is
also predominantly neutral, with High-PC3 yielding two improvements, ten
ties, and three declines and Low-PC3 yielding four improvements, nine
ties, and two declines. The Chinese results therefore identify Low-PC2 as
the most promising aggregate direction, but not as a transferable
cross-model rule.

\paragraph{Magnitude of Change.}
The directional analysis measures how often PCA-guided subsets outperform Random, whereas the normalized margin captures the average size of the performance difference; the two are therefore interpreted jointly.
Table~\ref{tab:margin_extremes} summarizes the largest positive and negative dataset-level normalized margins for each language and model.
In Arabic, PC1 produces some of the largest gains for LLaMA and Mistral but substantial declines for DeepSeek, making it influential but strongly architecture dependent. PC2 is generally weak or negative. PC3 provides the most reliable Arabic pattern overall: it is broadly favourable for LLaMA and DeepSeek and produces several positive Mistral effects, although these are offset by a large decline for Aya dataset.

Japanese effects are more source dependent. High-PC1 is the clearest positive pattern for LLaMA, while PC2 and PC3 produce larger gains only for particular model--dataset combinations; notably, PC3 yields substantial
margins despite its tie-dominated directional profile. Chinese results are similarly model specific: Low-PC1 is most favourable for LLaMA, Low-PC2 and PC3 for DeepSeek, and PC3 for Mistral. Thus, PCA-guided selection can produce meaningful gains, but the preferred component and direction depend on both the model and source dataset. Chinese results remain exploratory because of the smaller available subsets.

\section{Discussion}
\label{sec:discussion}

Our results show that lightweight linguistic, semantic, and structural
features provide useful descriptions of post-training datasets before
fine-tuning. PCA compresses these features into interpretable, language-specific dimensions, allowing datasets to be compared according to their linguistic properties before fine-tuning. However, the
correlation analysis also shows that no principal component is a consistent
predictor of cultural performance: the relevant component and its
direction vary across languages, benchmarks, and model architectures.

The controlled interventions further show that these descriptors can be
actionable, but that directional recurrence and effect magnitude must be
interpreted jointly. Improvement--Tie--Decline counts indicate how
consistently a subset outperforms Random across benchmarks, whereas
normalized margins capture the average size of the performance difference.
Agreement between the two provides the clearest evidence for a useful
selection signal; disagreement distinguishes recurrent small effects from
less frequent but larger changes.

The resulting selection patterns are both language and architecture
dependent. Arabic shows the most recurrent benefits from PC3-guided
selection, while PC1 can produce larger but less stable effects. Japanese
shows the clearest recurrent pattern under High-PC1, particularly for
LLaMA, whereas PC3 produces gains for specific model--dataset combinations
rather than a general filtering rule. Chinese results are weaker and
exploratory, with promising directions differing across architectures,
including Low-PC2 and PC3 for DeepSeek, Low-PC1 for LLaMA, and PC3 for
Mistral.

Importantly, these differences indicate that post-training data should not
be assumed to have the same value for every model family. The same
linguistic selection direction can improve one architecture while
degrading another, suggesting that dataset quality for cultural alignment
is partly model dependent. Accordingly, High-PC and Low-PC subsets are best understood as opposite ends of an interpretable linguistic dimension, rather than as intrinsically good and bad data.

In practice, the framework can be used to compare candidate subsets before training and identify the linguistic directions most promising for cultural fine-tuning in a given language and model. Table~\ref{tab:practical_takeaways} summarizes the main PCA-guided selection patterns observed across the three languages. This can reduce unnecessary fine-tuning runs and focus evaluation on the most informative data configurations rather than treating all candidate subsets as equally promising.

\begin{table}[t]
\centering
\begin{tabular}{lp{0.66\linewidth}}
\toprule
\textbf{Language} & \textbf{Main finding} \\
\midrule
Arabic &
PC3 is the most reliable signal; PC1 has larger but less stable effects. \\
Japanese &
High-PC1 is the clearest recurrent pattern, especially for LLaMA; PC3
produces dataset-specific gains. \\
Chinese &
Results are exploratory and architecture dependent, with the preferred
component and direction varying across models. \\
\bottomrule
\end{tabular}
\caption{Summary of PCA-guided selection patterns.}
\label{tab:practical_takeaways}
\end{table}
\section{Conclusion}

We introduced a dataset-centric framework for examining how linguistic,
semantic, and structural properties of post-training data relate to the
cultural performance of fine-tuned LLMs. The framework characterizes
datasets before training, reduces their properties into language-specific
PCA dimensions, and evaluates these dimensions through correlations and
controlled High-PC, Low-PC, and Random subset interventions.
Across Arabic, Japanese, and Chinese, dataset structure is associated with
cultural performance, but the relevant component and selection direction
vary across languages, benchmarks, source datasets, and model families. In particular, a selection direction that benefits one architecture may degrade another, showing that post-training data quality for cultural alignment is model dependent.
PCA-guided selection should therefore be used as a language- and
architecture-aware diagnostic rather than as a universal ranking of
dataset quality. By using lightweight dataset measurements to identify
promising selection directions before fine-tuning, the framework offers a
more cost-effective approach to culturally aligned post-training-data
design.
\section{Limitations and Future Work}

The correlation analysis is based on only 9--13 datasets per language,
which limits statistical power and produces wide confidence intervals for
some estimates. We therefore interpret the correlations as descriptive
associations rather than definitive evidence of general or causal effects.
Dataset features were computed from fixed random samples of 1,000 examples;
although pilot checks with larger samples and repeated sampling produced
similar feature values and PCA structure, some sampling variability may
remain. Evaluation used a standardized single-response setting and each
benchmark's designated scoring method. Different decoding or aggregation
strategies could change absolute scores, although our analysis focuses on
relative differences under a common setup. Finally, the experiments cover
three languages and three model families. Broader validation across
additional languages, datasets, architectures, and larger subset pools is
needed to determine how widely the observed model- and language-specific
selection patterns generalize.

\bibliography{custom}
\clearpage

\appendix

\section{Training Dataset Statistic}\label{appendix:dataset_stats}

We summarize the training datasets used in our study in Table~\ref{tab:dataset_stats}. The datasets span three languages (Arabic, Japanese, and Chinese) and cover diverse domains. They are collected from a variety of sources such as benchmark datasets, web data, media articles, and social media platforms, ensuring broad coverage of linguistic styles and cultural contexts.


\begin{table*}[htpb]
\centering
\footnotesize
\setlength{\tabcolsep}{4pt}
\begin{tabular}{p{4.9cm} c c p{3cm} p{3cm}}
\toprule
\textbf{Dataset} & \textbf{Rows} & \textbf{Lang} & \textbf{Domain} & \textbf{Source Type} \\
\midrule

MultiNativQA~\citep{hasan2024nativqa} & $\sim$5k & AR & Question Answering / Cultural & Crowdsourced \\
ArabicMMLU~\citep{koto2024arabicmmlu} & $\sim$15k & AR & Academic / Knowledge & Benchmark Dataset \\
Aya~\citep{singh2024aya}$^{*}$ & $\sim$5k & AR & Instruction / General & Synthetic and Curated \\
CIDAR~\citep{alyafeai2024cidar} & 10k & AR & Cultural Alignment & Mixed Sources \\
Open-ArabicaQA~\citep{abdallah2024arabicaqa} & $\sim$62k & AR & Question Answering & Web Data \\
The Ultimate Arabic News~\citep{ultimatenews} & 196k & AR & News & Media Articles \\
DAWQAS~\citep{ismail2018dawqas} & $\sim$3k & AR & Question Answering & Web Data \\
Al Jazeera News Articles~\citep{arbml_aljazeera_news_articles} & 6k & AR & News & Media Articles \\
Arsen-20~\citep{arbml_arsen20} & 20k & AR & Sentiment / Social Media & Social Media Data \\
Ichikara Instruction All~\citep{sekine2024ichikara} & 10k & JP & Instruction Tuning & Crowdsourced \\
Databricks-Dolly-JA~\citep{kunishou_dolly15k_ja} & 15k & JP & Instruction Tuning & Machine-translated (from Dolly-15k) \\
JaQuAD~\citep{so2022jaquad} & $\sim$32k & JP & Question Answering & Crowdsourced (Wikipedia-based) \\
WordNet 2.0~\citep{bond-kuribayashi-2023-japanese} & 158k & JP & Lexical Resource & Curated Lexical Database \\
Jcommonsense~\citep{takeshita2023jcommonsense} & 9k & JP & Ethics / Commonsense Reasoning & Crowdsourced (ConceptNet-based)\\

AIO Instruction Dataset~\citep{aio_dataset} & 23k & JP & Instruction Tuning & Mixed Sources (competition, Wikipedia) \\
Easy-Japanese~\citep{katsuta2018simplification} & 34k & JP & Text Simplification & Crowdsourced (textbook-based, curated) \\
Wiki JP~\citep{wikipedia_ja_20230720, wikimedia_dumps} & 30k & JP & General Knowledge & Encyclopedia Data \\
Wikinews JP~\citep{Hirano2023-llmj,wikinews_japanese} & 5k & JP & News & Media Articles \\
Aozora Bunko~\citep{aozora_txt} & 30k & JP & Literature & Books and Literary Texts \\

COIG-PC~\citep{bai2024coig} & 3k & ZH & Instruction / General & Mixed Sources \\
Chinese\_Traditional~\citep{bai2024coig} & 1k & ZH & General Domain & Web Data \\
Douban~\citep{bai2024coig} & 3k & ZH & Social Media / Reviews & Social Media Data \\
Xiaohongshu (XHS)~\citep{bai2024coig} & $\sim$2k & ZH & Social Media / Lifestyle & Social Media Data \\
Human\_Value~\citep{bai2024coig} & 1k & ZH & Cultural Values & Mixed Sources \\
RuozhiBa~\citep{bai2024coig} & $\sim$0.25k & ZH & Social Media / Humor & Social Media Data \\
Exam~\citep{bai2024coig} & 5k & ZH & Academic & Benchmark Dataset \\
SegmentFault~\citep{bai2024coig} & 0.5k & ZH & Technical Discussions & Online Forum \\
Zhihu~\citep{bai2024coig} & $\sim$6k & ZH & Question Answering / Discussion & Online Forum \\
LogiQA~\citep{bai2024coig} & $\sim$0.5k & ZH & Logical Reasoning & Benchmark Dataset \\
Wikihow~\citep{bai2024coig} & $\sim$1.5k & ZH & Instructional Content & Web Data \\
Wiki~\citep{bai2024coig} & $\sim$11k & ZH & General Knowledge & Encyclopedia Data \\
Finance~\citep{bai2024coig} & $\sim$11k & ZH & Financial Domain & Web Data \\

\bottomrule
\end{tabular}
\caption{Summary statistics of datasets, including size, language, domain, and source type.}
\centering
{\footnotesize $^{*}$ Arabic subset obtained via language filtering.}
\label{tab:dataset_stats}
\end{table*}

\section{Benchmarks by Language}
\label{appendix:benchmarks}

Table~\ref{tab:lang_benchmarks} lists the cultural and value-alignment benchmarks used for evaluation in each language. These benchmarks span multiple dimensions of cultural knowledge, norms, and moral reasoning.

\begin{table*}[htpb]
\centering
\small
\begin{tabular}{l p{3.65cm} p{3.65cm} p{3.649cm}}
\toprule
\textbf{Language} & \textbf{Cultural Knowledge} & \textbf{Cultural Values} & \textbf{Cultural Norms} \\
\midrule
Arabic &
CulturalBench (Easy, Hard)~\citep{chiu2024culturalbench}, 
CultureAtlas~\citep{fung2024massively}, 
ArabCulture~\citep{sadallah2025arabculture}, 
EXAMs-AR~\citep{sengupta2023jais} &
VSM13 (Arabic)~\citep{hofstede2013vsm}, 
WorldValuesBench~\citep{zhao2024worldvaluesbench}, 
ACVA-Arabic~\citep{huang2023acegpt} &
CIDAR-MCQ, CIDAR-EVAL~\citep{alyafeai2024cidar}, 
AraDiCE~\citep{mousi2024aradice}, 
LLM-Globe (Open, Closed)~\citep{karinshak2024llm} \\
\midrule
Japanese &
CulturalBench (Easy, Hard)~\citep{chiu2024culturalbench} &
VSM13~\citep{hofstede2013vsm}, 
WorldValuesBench~\citep{zhao2024worldvaluesbench} &
J-Ethics~\cite{takeshita2025jethicsjapaneseethicsunderstanding}, 
LLM-Globe (Open, Closed)~\citep{karinshak2024llm} \\
\midrule
Chinese &
CulturalBench (Easy, Hard)~\cite{chiu2024culturalbench}, 
CultureAtlas~\citep{fung2024massively}, 
CEval-Exams~\citep{huang2023ceval} &
VSM13~\citep{hofstede2013vsm}, 
WorldValuesBench~\citep{zhao2024worldvaluesbench} &
CMoral~\citep{yu2024cmoralevalmoralevaluationbenchmark}, 
LLM-Globe (Open, Closed)~\citep{karinshak2024llm} \\
\bottomrule
\end{tabular}
\caption{Cultural alignment benchmarks by category and language.}\label{tab:lang_benchmarks} 
\end{table*}

\section{Additional Correlation Results}\label{appendix:additional_results}

This section provides detailed correlation results between dataset linguistic
properties (PCA components PC1--PC3) and downstream cultural evaluation
metrics across languages and models. Results for LLaMA, Mistral, and DeepSeek
are reported for Arabic, Japanese, and Chinese in
Tables~\ref{tab:arabic_correlations}--\ref{tab:chinese_correlations}. Correlations are presented with 95\% confidence intervals to reflect variability due to the limited number of datasets. These results complement the main analysis by providing a more fine-grained view of how different dataset characteristics relate to cultural alignment performance.

Across all three languages, benchmark performance shows non-zero
correlations with at least one PCA dimension. However, the component most
strongly associated with performance varies across languages, model
architectures, and benchmark categories. No single PCA component therefore
consistently predicts higher or lower cultural performance across all
experimental settings.

\subsection{Arabic}

For Arabic (Table~\ref{tab:arabic_correlations}), associations appear across all
benchmark categories but differ by model architecture. For example, the
cultural knowledge benchmark CultureAtlas correlates most strongly with PC1
for LLaMA ($0.85$), whereas EXAMs correlates with PC2 for Mistral ($0.82$).
Cultural values benchmarks also shift across components, with
WorldValuesBench correlating with PC1 for Mistral ($-0.77$) and ACVA with
PC2 for DeepSeek ($0.78$). Normative benchmarks similarly vary across
models: CIDAR-EVAL correlates with PC2 for LLaMA ($0.71$) and with PC3 for
DeepSeek ($0.60$). These results show that the PCA dimension most strongly
associated with cultural performance depends on both the benchmark and the
model architecture.

\begin{table*}[htbp]
\centering
\scriptsize
\setlength{\tabcolsep}{4pt}

\begin{tabular}{llccc}
\toprule
\textbf{PC} & \textbf{Benchmark}
& \textbf{LLaMA}
& \textbf{Mistral}
& \textbf{DeepSeek} \\
\midrule

\multicolumn{5}{c}{\textbf{PC1}} \\
\midrule
PC1 & ACVA
& 0.46 [-0.74, 0.89]
& --
& 0.21 [-0.65, 0.70] \\

PC1 & AraDiCE-Culture
& -0.38 [-0.93, 0.38]
& -0.59 [-0.93, -0.05]
& 0.29 [-0.42, 0.81] \\

PC1 & ArabCulture
& -0.60 [-0.94, -0.16]
& 0.02 [-0.74, 0.78]
& -0.21 [-0.96, 0.62] \\

PC1 & CIDAR-EVAL (GPT-Judge)
& 0.05 [-0.57, 0.84]
& -0.51 [-0.90, 0.15]
& -0.04 [-0.66, 0.56] \\

PC1 & CIDAR-MCQ
& -0.33 [-0.83, 0.74]
& 0.03 [-0.73, 0.70]
& 0.74 [0.12, 0.99] \\

PC1 & CulturalBench-Easy
& -0.51 [-0.89, 0.36]
& -0.67 [-0.90, -0.12]
& 0.30 [-0.33, 0.80] \\

PC1 & CulturalBench-Hard
& -0.64 [-0.93, -0.06]
& -0.15 [-0.92, 0.68]
& 0.09 [-0.67, 0.76] \\

PC1 & CultureAtlas
& 0.85 [0.75, 0.97]
& -0.71 [-0.94, -0.27]
& 0.13 [-0.69, 0.88] \\

PC1 & EXAMs
& 0.19 [-0.71, 0.77]
& 0.17 [-0.67, 0.93]
& -- \\

PC1 & LLM-GLOBE (Closed)
& 0.38 [-0.60, 0.87]
& -0.55 [-0.90, 0.07]
& 0.37 [-0.47, 0.83] \\

PC1 & LLM-GLOBE (Open)
& -0.01 [-0.71, 0.62]
& -0.58 [-0.88, -0.23]
& 0.18 [-0.55, 0.78] \\

PC1 & VSM13
& -0.41 [-0.84, 0.24]
& 0.15 [-0.61, 0.75]
& 0.33 [-0.56, 0.84] \\

PC1 & WorldValuesBench (EMD)
& -0.20 [-0.88, 0.80]
& -0.77 [-0.94, -0.45]
& -0.30 [-0.74, 0.59] \\

\midrule
\multicolumn{5}{c}{\textbf{PC2}} \\
\midrule

PC2 & ACVA
& 0.16 [-0.38, 0.73]
& --
& 0.78 [0.41, 0.97] \\

PC2 & AraDiCE-Culture
& -0.36 [-0.91, 0.42]
& 0.05 [-0.82, 0.87]
& -0.26 [-0.79, 0.55] \\

PC2 & ArabCulture
& -0.51 [-0.86, -0.02]
& -0.41 [-0.92, 0.29]
& -0.48 [-0.94, 0.21] \\

PC2 & CIDAR-EVAL (GPT-Judge)
& 0.71 [0.14, 0.96]
& -0.10 [-0.95, 0.55]
& 0.19 [-0.56, 0.77] \\

PC2 & CIDAR-MCQ
& 0.50 [-0.22, 0.91]
& -0.35 [-0.96, 0.49]
& 0.12 [-0.50, 0.77] \\

PC2 & CulturalBench-Easy
& 0.18 [-0.87, 0.76]
& 0.29 [-0.54, 0.85]
& 0.26 [-0.58, 0.89] \\

PC2 & CulturalBench-Hard
& 0.43 [-0.54, 0.97]
& 0.27 [-0.46, 0.85]
& 0.30 [-0.40, 0.84] \\

PC2 & CultureAtlas
& 0.06 [-0.57, 0.71]
& -0.27 [-0.87, 0.47]
& -0.08 [-0.77, 0.79] \\

PC2 & EXAMs
& -0.39 [-0.94, 0.42]
& 0.82 [0.49, 0.96]
& -- \\

PC2 & LLM-GLOBE (Closed)
& -0.24 [-0.80, 0.29]
& -0.41 [-0.87, -0.12]
& -0.71 [-0.92, -0.57] \\

PC2 & LLM-GLOBE (Open)
& -0.53 [-0.91, 0.39]
& 0.09 [-0.92, 0.74]
& -0.56 [-0.98, 0.15] \\

PC2 & VSM13
& 0.38 [-0.35, 0.89]
& 0.15 [-0.69, 0.83]
& 0.09 [-0.60, 0.68] \\

PC2 & WorldValuesBench (EMD)
& 0.51 [-0.24, 0.84]
& 0.02 [-0.68, 0.78]
& -0.60 [-0.89, -0.08] \\

\midrule
\multicolumn{5}{c}{\textbf{PC3}} \\
\midrule

PC3 & ACVA
& -0.35 [-0.88, 0.10]
& --
& 0.41 [-0.53, 0.87] \\

PC3 & AraDiCE-Culture
& 0.36 [-0.38, 0.95]
& -0.55 [-0.96, 0.32]
& 0.04 [-0.84, 0.76] \\

PC3 & ArabCulture
& -0.06 [-0.75, 0.72]
& 0.23 [-0.33, 0.71]
& 0.09 [-0.41, 0.63] \\

PC3 & CIDAR-EVAL (GPT-Judge)
& 0.41 [-0.19, 0.82]
& -0.01 [-0.73, 0.62]
& 0.60 [-0.21, 0.96] \\

PC3 & CIDAR-MCQ
& 0.10 [-0.68, 0.75]
& -0.26 [-0.89, 0.57]
& -0.55 [-0.88, -0.18] \\

PC3 & CulturalBench-Easy
& 0.40 [0.03, 0.84]
& -0.23 [-0.74, 0.57]
& 0.14 [-0.88, 0.81] \\

PC3 & CulturalBench-Hard
& -0.33 [-0.95, 0.59]
& 0.10 [-0.43, 0.78]
& -0.11 [-0.75, 0.62] \\

PC3 & CultureAtlas
& 0.11 [-0.65, 0.84]
& -0.13 [-0.80, 0.78]
& 0.52 [0.04, 0.95] \\

PC3 & EXAMs
& 0.16 [-0.67, 0.76]
& 0.15 [-0.61, 0.76]
& -- \\

PC3 & LLM-GLOBE (Closed)
& 0.29 [-0.38, 0.96]
& 0.13 [-0.66, 0.82]
& -0.43 [-0.88, -0.10] \\

PC3 & LLM-GLOBE (Open)
& -0.56 [-0.87, -0.04]
& -0.54 [-0.83, 0.23]
& 0.68 [0.02, 0.98] \\

PC3 & VSM13
& -0.57 [-0.90, 0.08]
& -0.47 [-0.89, 0.17]
& -0.22 [-0.73, 0.33] \\

PC3 & WorldValuesBench (EMD)
& 0.09 [-0.54, 0.59]
& -0.19 [-0.69, 0.52]
& -0.46 [-0.87, 0.27] \\

\bottomrule
\end{tabular}

\caption{Correlations between dataset PCA components (PC1--PC3) and
downstream cultural performance for Arabic across model families.
Values report Pearson's \(r\) with 95\% confidence intervals.}
\label{tab:arabic_correlations}
\end{table*}

\subsection{Japanese}

For Japanese (Table~\ref{tab:japanese_correlations}), cultural knowledge benchmarks
most frequently align with PC1. For CulturalBench-Easy, the PC1 correlations
are $-0.66$ for LLaMA, $-0.68$ for Mistral, and $-0.43$ for DeepSeek.
In contrast, cultural values and norms benchmarks shift across PC1, PC2,
and PC3 depending on the model. For example, VSM13 correlates most strongly
with PC2 for LLaMA ($0.73$) but with PC1 for Mistral ($0.74$), while
JETHICS and LLM-GLOBE (Closed) correlate with PC3 for Mistral ($0.55$ and
$-0.54$, respectively). These results indicate architecture-dependent
sensitivity to different dimensions of Japanese dataset structure.

\begin{table*}[htbp]
\centering
\scriptsize
\setlength{\tabcolsep}{4pt}

\begin{tabular}{llccc}
\toprule
\textbf{PC} & \textbf{Benchmark}
& \textbf{LLaMA}
& \textbf{Mistral}
& \textbf{DeepSeek} \\
\midrule

\multicolumn{5}{c}{\textbf{PC1}} \\
\midrule

PC1 & CulturalBench-Easy
& -0.66 [-0.96, 0.25]
& -0.68 [-0.94, -0.04]
& -0.43 [-0.88, 0.06] \\

PC1 & CulturalBench-Hard
& -0.36 [-0.81, 0.04]
& -0.26 [-0.80, 0.55]
& -0.32 [-0.78, 0.23] \\

PC1 & JETHICS
& -0.18 [-0.83, 0.31]
& -0.53 [-0.83, 0.15]
& 0.53 [-0.00, 0.99] \\

PC1 & LLM-GLOBE (Closed)
& 0.34 [-0.12, 0.73]
& -0.11 [-0.79, 0.80]
& 0.04 [-0.49, 0.78] \\

PC1 & LLM-GLOBE (Open)
& -0.18 [-0.83, 0.53]
& 0.41 [-0.41, 0.84]
& 0.13 [-0.48, 0.62] \\

PC1 & VSM13
& 0.19 [-0.38, 0.75]
& 0.74 [0.23, 0.95]
& 0.19 [-0.55, 0.71] \\

PC1 & WorldValuesBench (EMD)
& 0.34 [-0.35, 0.82]
& 0.32 [-0.25, 0.79]
& -0.36 [-0.89, 0.27] \\

\midrule
\multicolumn{5}{c}{\textbf{PC2}} \\
\midrule

PC2 & CulturalBench-Easy
& 0.50 [0.04, 0.83]
& -0.28 [-0.78, 0.33]
& -0.10 [-0.52, 0.34] \\

PC2 & CulturalBench-Hard
& -0.05 [-0.51, 0.73]
& 0.16 [-0.23, 0.71]
& 0.24 [-0.54, 0.93] \\

PC2 & JETHICS
& 0.12 [-0.67, 0.67]
& -0.01 [-0.63, 0.67]
& 0.12 [-0.74, 0.87] \\

PC2 & LLM-GLOBE (Closed)
& -0.21 [-0.85, 0.30]
& 0.45 [-0.64, 0.89]
& 0.54 [0.16, 0.92] \\

PC2 & LLM-GLOBE (Open)
& -0.21 [-0.77, 0.44]
& -0.34 [-0.88, 0.30]
& 0.32 [-0.51, 0.95] \\

PC2 & VSM13
& 0.73 [0.17, 0.95]
& -0.09 [-0.86, 0.79]
& 0.35 [-0.56, 0.82] \\

PC2 & WorldValuesBench (EMD)
& -0.06 [-0.44, 0.33]
& -0.08 [-0.75, 0.56]
& -0.70 [-0.99, 0.59] \\

\midrule
\multicolumn{5}{c}{\textbf{PC3}} \\
\midrule

PC3 & CulturalBench-Easy
& 0.02 [-0.62, 0.83]
& 0.23 [-0.39, 0.79]
& 0.14 [-0.41, 0.77] \\

PC3 & CulturalBench-Hard
& -0.27 [-0.74, 0.55]
& 0.46 [-0.14, 0.83]
& 0.26 [-0.42, 0.88] \\

PC3 & JETHICS
& -0.22 [-0.66, 0.35]
& 0.55 [0.24, 0.89]
& -0.09 [-0.81, 0.64] \\

PC3 & LLM-GLOBE (Closed)
& -0.32 [-0.73, 0.13]
& -0.54 [-0.89, -0.14]
& 0.21 [-0.47, 0.72] \\

PC3 & LLM-GLOBE (Open)
& -0.40 [-0.91, 0.26]
& -0.25 [-0.76, 0.67]
& -0.23 [-0.75, 0.47] \\

PC3 & VSM13
& 0.06 [-0.62, 0.81]
& 0.26 [-0.54, 0.86]
& -0.30 [-0.93, 0.46] \\

PC3 & WorldValuesBench (EMD)
& -0.47 [-0.88, 0.39]
& -0.05 [-0.57, 0.66]
& -0.01 [-0.72, 0.69] \\

\bottomrule
\end{tabular}

\caption{Correlations between dataset PCA components (PC1--PC3) and
downstream cultural performance for Japanese across model families.
Values report Pearson's \(r\) with 95\% confidence intervals.}
\label{tab:japanese_correlations}
\end{table*}

\subsection{Chinese}

For Chinese (Table~\ref{tab:chinese_correlations}), correlations are distributed
across components with no consistent dominant axis. For LLaMA,
LLM-GLOBE (Open) correlates with PC1 ($-0.50$), CulturalBench-Hard with
PC2 ($0.40$), and CMoralEval with PC3 ($0.62$). For Mistral, several of
the strongest associations occur on PC1, including CulturalBench-Easy
($-0.37$), LLM-GLOBE (Closed) ($0.77$), and VSM13 ($0.51$), while
CultureAtlas and C-Eval correlate with PC3 ($-0.34$ and $-0.45$,
respectively). For DeepSeek, WorldValuesBench correlates with both PC1
($0.69$) and PC2 ($0.43$), whereas CultureAtlas correlates with PC3
($-0.58$). These results show that the relevant component varies across
models and benchmarks.

\begin{table*}[htbp]
\centering
\scriptsize
\setlength{\tabcolsep}{4pt}

\begin{tabular}{llccc}
\toprule
\textbf{PC} & \textbf{Benchmark}
& \textbf{LLaMA}
& \textbf{Mistral}
& \textbf{DeepSeek} \\
\midrule

\multicolumn{5}{c}{\textbf{PC1}} \\
\midrule

PC1 & CMoralEval
& 0.11 [-0.48, 0.58]
& -0.28 [-0.78, 0.32]
& -0.60 [-0.88, 0.25] \\

PC1 & CulturalBench-Easy
& -0.18 [-0.77, 0.16]
& -0.37 [-0.65, -0.04]
& 0.19 [-0.13, 0.55] \\

PC1 & CulturalBench-Hard
& -0.02 [-0.62, 0.83]
& 0.36 [-0.03, 0.72]
& -0.10 [-0.65, 0.24] \\

PC1 & CultureAtlas
& 0.15 [-0.31, 0.71]
& -0.25 [-0.67, 0.45]
& 0.32 [-0.11, 0.85] \\

PC1 & C-Eval
& -0.15 [-0.50, 0.31]
& 0.15 [-0.25, 0.69]
& 0.09 [-0.39, 0.46] \\

PC1 & LLM-GLOBE (Closed)
& 0.21 [-0.16, 0.52]
& 0.77 [0.28, 0.93]
& -- \\

PC1 & LLM-GLOBE (Open)
& -0.50 [-0.79, -0.06]
& -0.21 [-0.55, 0.32]
& -0.31 [-0.70, 0.06] \\

PC1 & VSM13
& 0.16 [-0.19, 0.65]
& 0.51 [0.06, 0.79]
& -0.41 [-0.73, 0.12] \\

PC1 & WorldValuesBench (EMD)
& 0.33 [-0.12, 0.67]
& -0.27 [-0.60, 0.15]
& 0.69 [0.11, 0.93] \\

\midrule
\multicolumn{5}{c}{\textbf{PC2}} \\
\midrule

PC2 & CMoralEval
& -0.06 [-0.46, 0.33]
& -0.11 [-0.61, 0.60]
& -0.31 [-0.87, 0.30] \\

PC2 & CulturalBench-Easy
& -0.20 [-0.76, 0.24]
& -0.23 [-0.62, 0.36]
& 0.51 [-0.32, 0.87] \\

PC2 & CulturalBench-Hard
& 0.40 [0.08, 0.77]
& 0.14 [-0.38, 0.83]
& -0.23 [-0.84, 0.57] \\

PC2 & CultureAtlas
& 0.21 [-0.56, 0.68]
& -0.36 [-0.80, 0.39]
& 0.31 [-0.19, 0.80] \\

PC2 & C-Eval
& 0.27 [-0.35, 0.90]
& -0.09 [-0.63, 0.46]
& -0.27 [-0.76, 0.36] \\

PC2 & LLM-GLOBE (Closed)
& -0.25 [-0.77, 0.19]
& -0.08 [-0.52, 0.38]
& -- \\

PC2 & LLM-GLOBE (Open)
& -0.39 [-0.82, 0.47]
& -0.18 [-0.70, 0.39]
& -0.45 [-0.84, 0.07] \\

PC2 & VSM13
& 0.45 [-0.18, 0.80]
& 0.10 [-0.45, 0.55]
& -0.13 [-0.51, 0.26] \\

PC2 & WorldValuesBench (EMD)
& 0.06 [-0.51, 0.45]
& -0.56 [-0.86, 0.05]
& 0.43 [0.03, 0.83] \\

\midrule
\multicolumn{5}{c}{\textbf{PC3}} \\
\midrule

PC3 & CMoralEval
& 0.62 [0.12, 0.86]
& -0.27 [-0.62, 0.14]
& -0.31 [-0.69, 0.39] \\

PC3 & CulturalBench-Easy
& 0.12 [-0.20, 0.76]
& 0.12 [-0.25, 0.48]
& -0.26 [-0.60, 0.03] \\

PC3 & CulturalBench-Hard
& -0.32 [-0.73, 0.17]
& -0.15 [-0.62, 0.39]
& 0.24 [-0.07, 0.68] \\

PC3 & CultureAtlas
& -0.20 [-0.61, 0.29]
& -0.34 [-0.77, -0.02]
& -0.58 [-0.83, -0.05] \\

PC3 & C-Eval
& -0.06 [-0.57, 0.42]
& -0.45 [-0.77, -0.12]
& -0.37 [-0.74, 0.14] \\

PC3 & LLM-GLOBE (Closed)
& -0.34 [-0.66, 0.01]
& -0.10 [-0.83, 0.66]
& -- \\

PC3 & LLM-GLOBE (Open)
& -0.02 [-0.44, 0.49]
& 0.09 [-0.52, 0.58]
& -0.06 [-0.34, 0.26] \\

PC3 & VSM13
& -0.19 [-0.69, 0.32]
& -0.18 [-0.71, 0.55]
& 0.42 [-0.45, 0.86] \\

PC3 & WorldValuesBench (EMD)
& 0.09 [-0.32, 0.55]
& -0.09 [-0.48, 0.33]
& 0.17 [-0.47, 0.68] \\

\bottomrule
\end{tabular}

\caption{Correlations between dataset PCA components (PC1--PC3) and
downstream cultural performance for Chinese across model families.
Values report Pearson's \(r\) with 95\% confidence intervals.}
\label{tab:chinese_correlations}
\end{table*}

Taken together, these results show that dataset linguistic structure is
associated with downstream cultural performance, but the relevant component
varies across models and benchmarks. The correlations therefore do not
provide a universal criterion for selecting a dataset dimension independently
of the target model. This motivates the controlled subset interventions
reported in the main paper, which test whether the PCA-derived dimensions
can be used directly to guide fine-tuning-data selection.

\section{Additional Correlation Heatmaps}
\label{appendix:correlation_heatmaps}

\begin{figure*}[htbp]
    \centering

    \begin{minipage}[t]{0.32\textwidth}
        \centering
        \includegraphics[width=\linewidth]{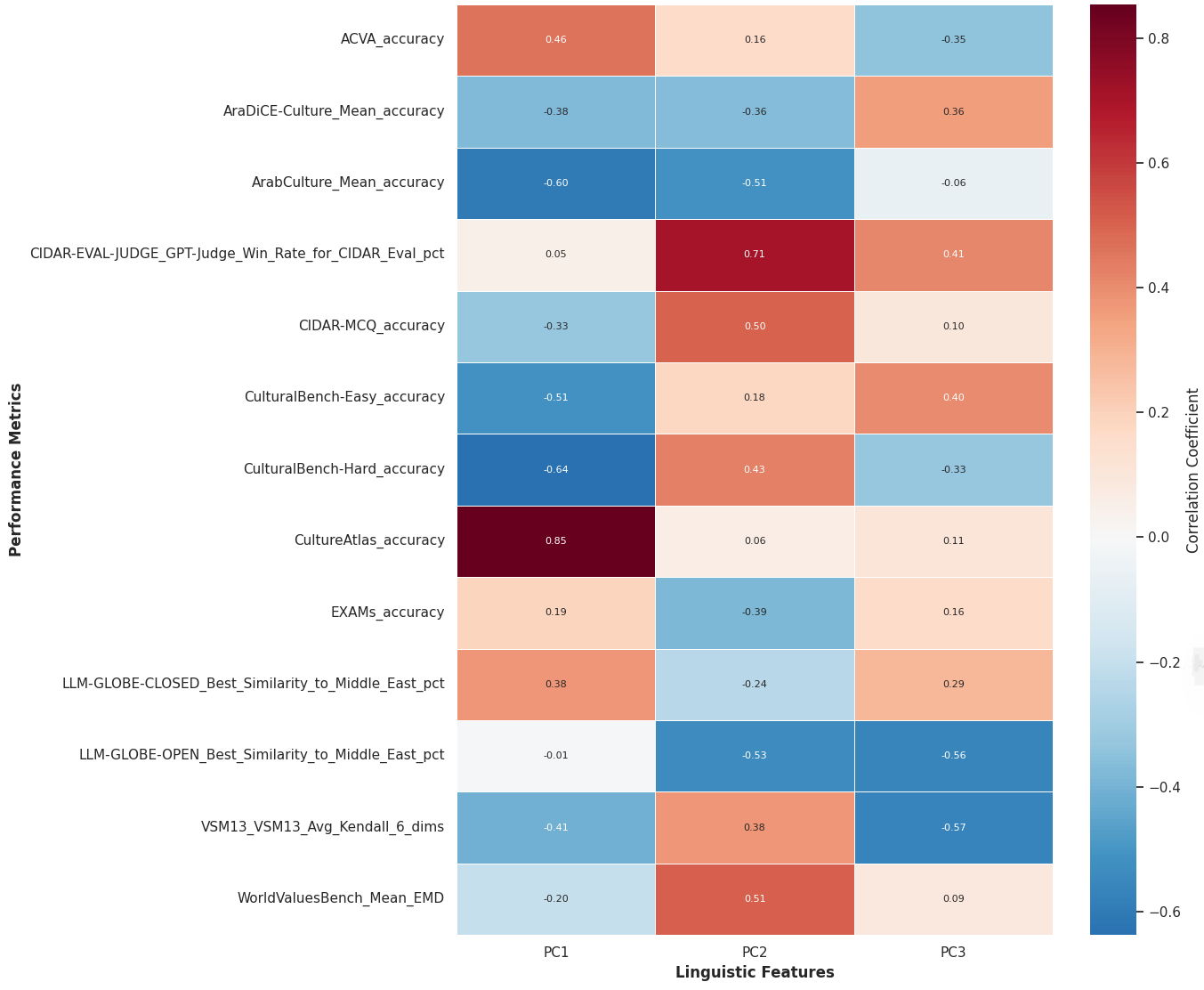}
        \caption*{(a) AR–LLaMA}
    \end{minipage}\hfill
    \begin{minipage}[t]{0.32\textwidth}
        \centering
        \includegraphics[width=\linewidth]{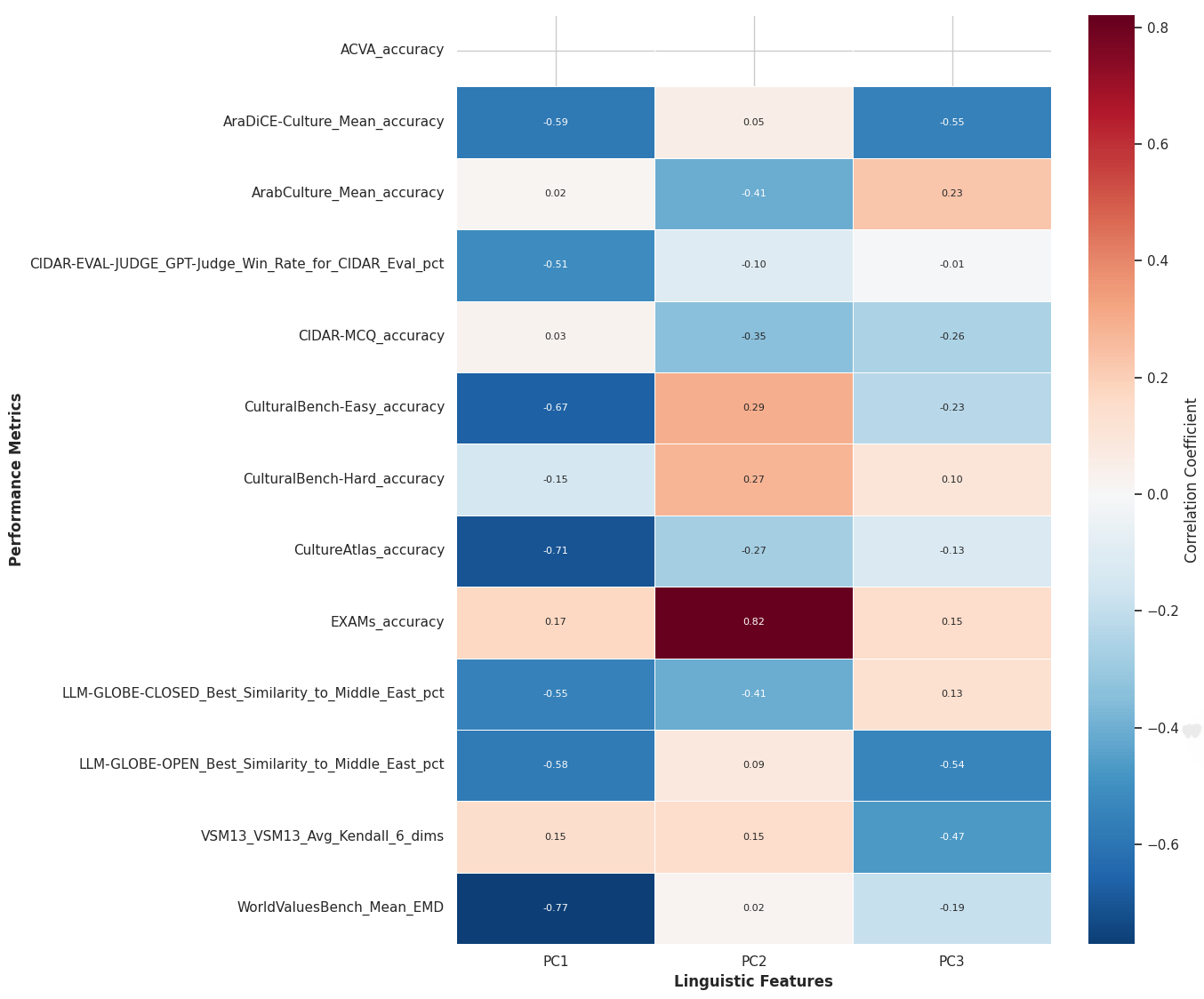}
        \caption*{(b) AR–Mistral}
    \end{minipage}\hfill
    \begin{minipage}[t]{0.32\textwidth}
        \centering
        \includegraphics[width=\linewidth]{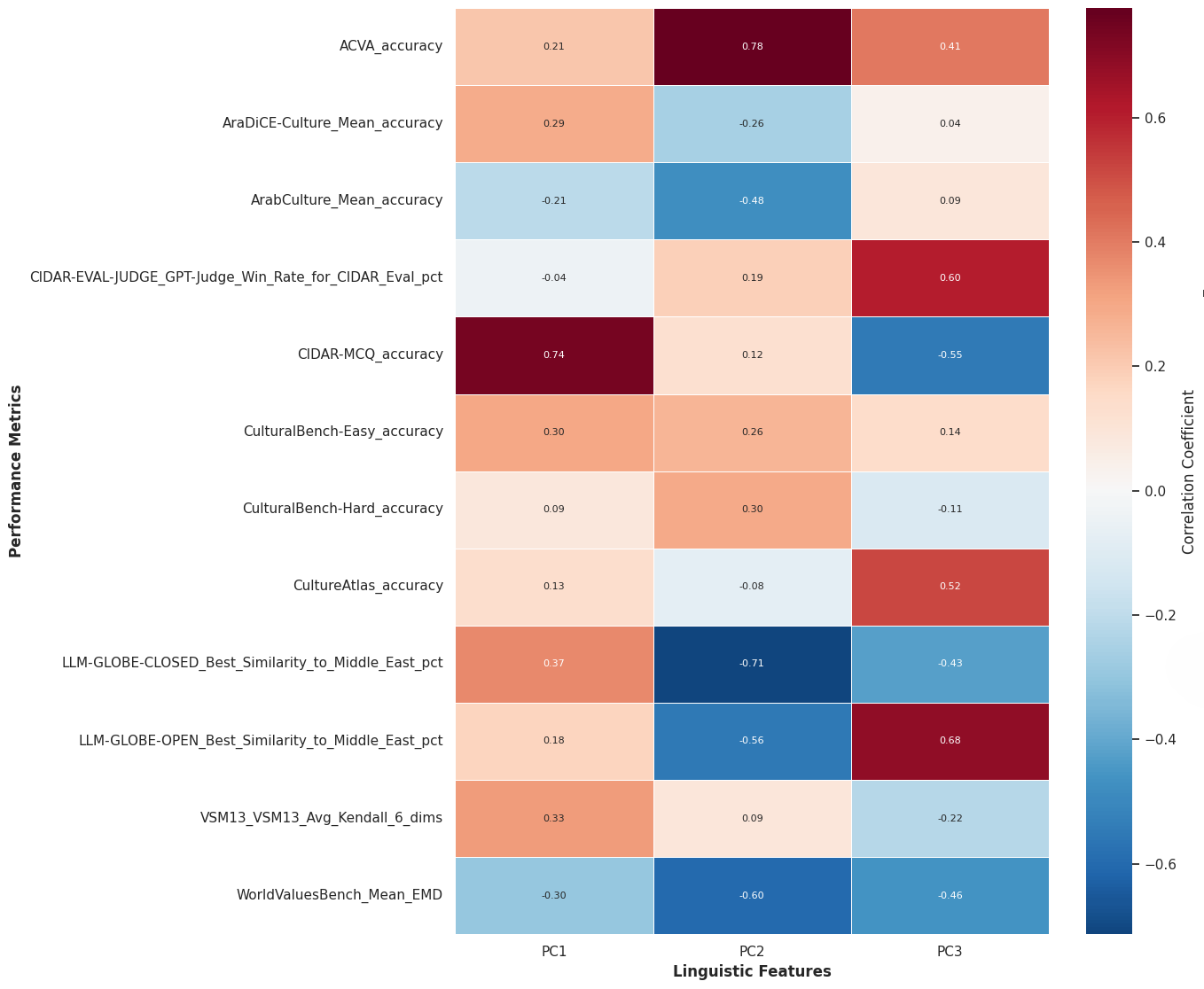}
        \caption*{(c) AR–DeepSeek}
    \end{minipage}
    \caption{Correlation between dataset PCA components (PC1--PC3) and downstream cultural performance for Arabic across models.}
    \label{fig:arabic_corr_heatmaps}
\end{figure*}

\begin{figure*}[htbp]
    \centering

    \begin{minipage}[t]{0.32\textwidth}
        \centering
        \includegraphics[width=\linewidth]{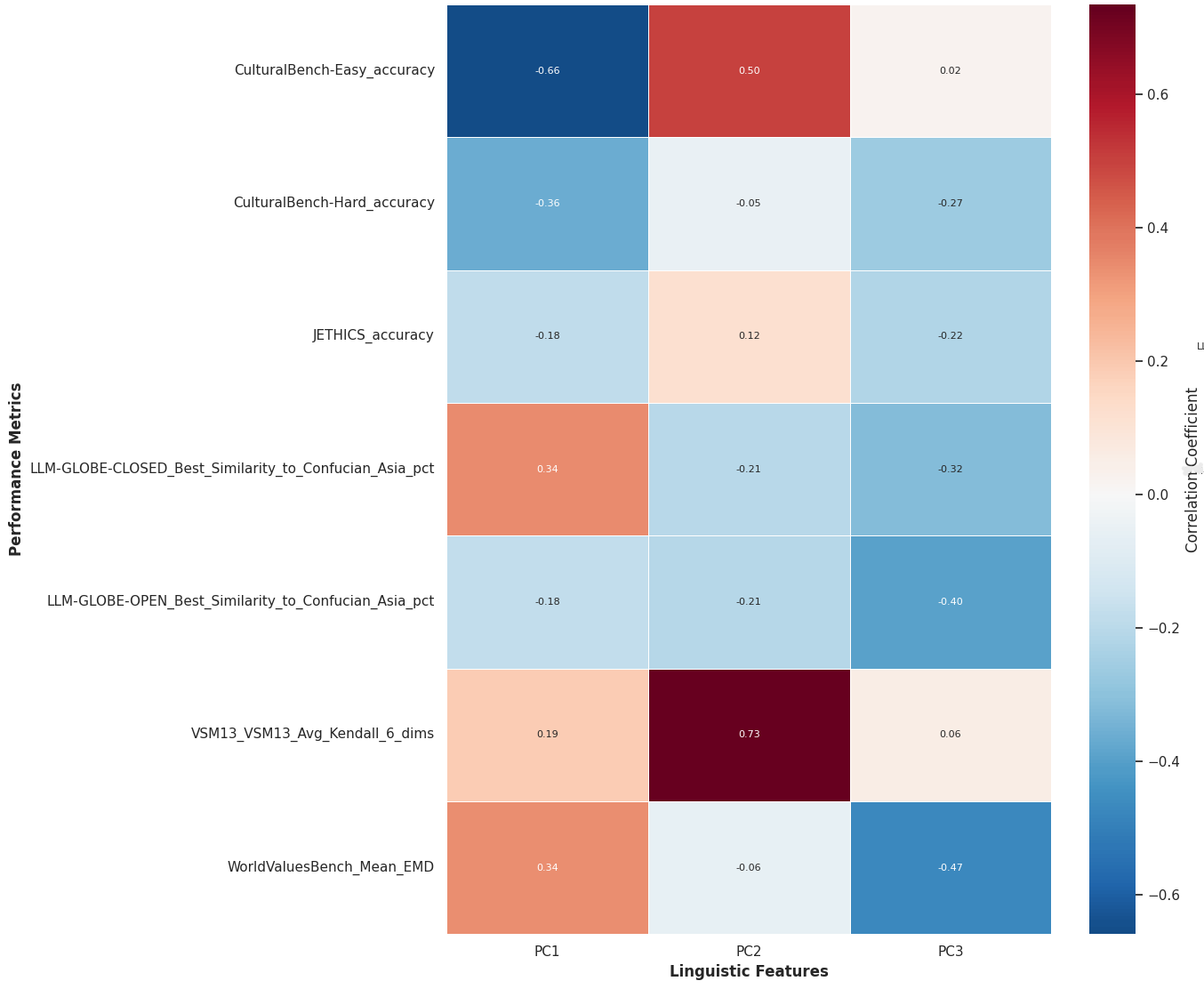}
        \caption*{(a) JP–LLaMA}
    \end{minipage}\hfill
    \begin{minipage}[t]{0.32\textwidth}
        \centering
        \includegraphics[width=\linewidth]{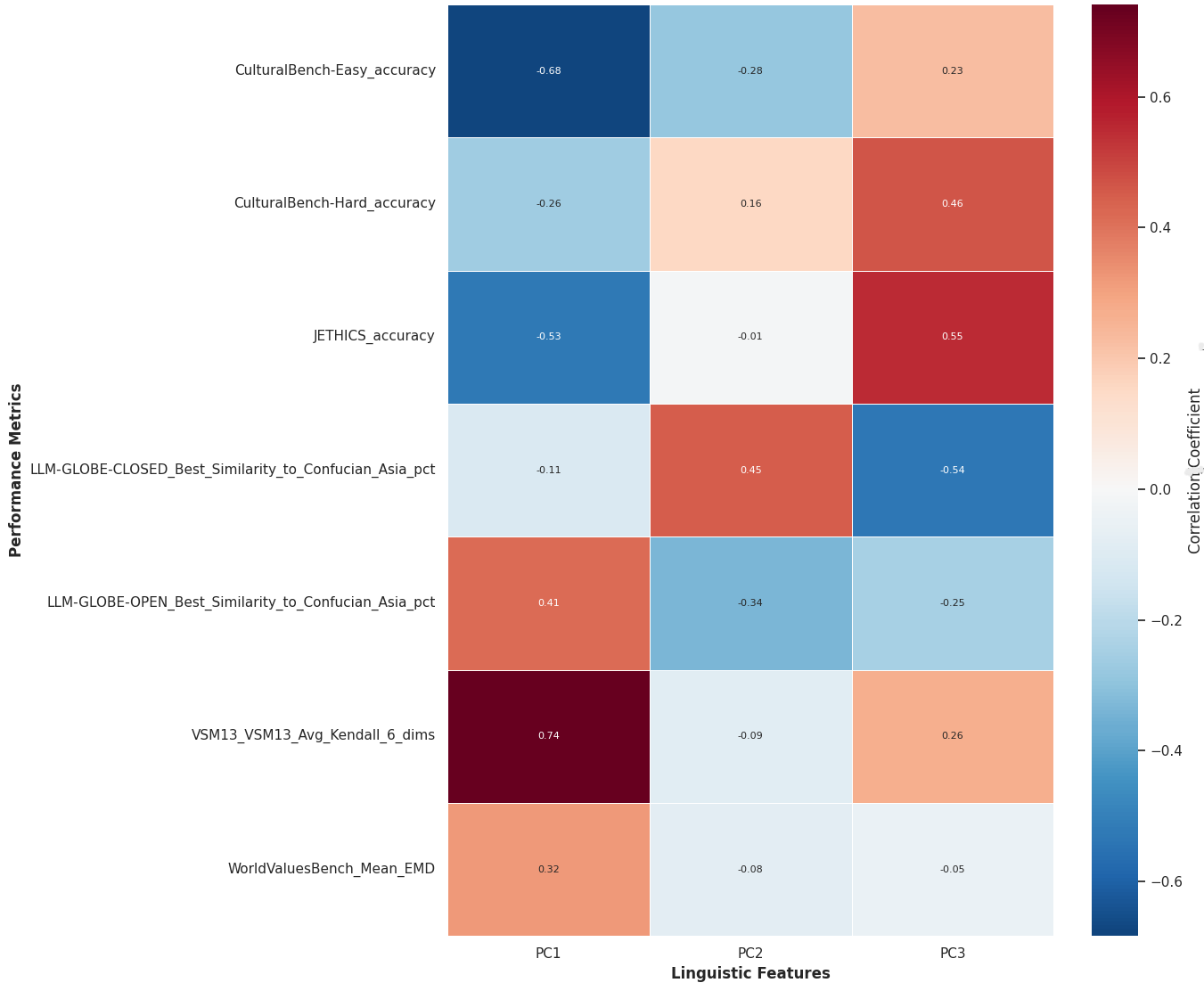}
        \caption*{(b) JP–Mistral}
    \end{minipage}\hfill
    \begin{minipage}[t]{0.32\textwidth}
        \centering
        \includegraphics[width=\linewidth]{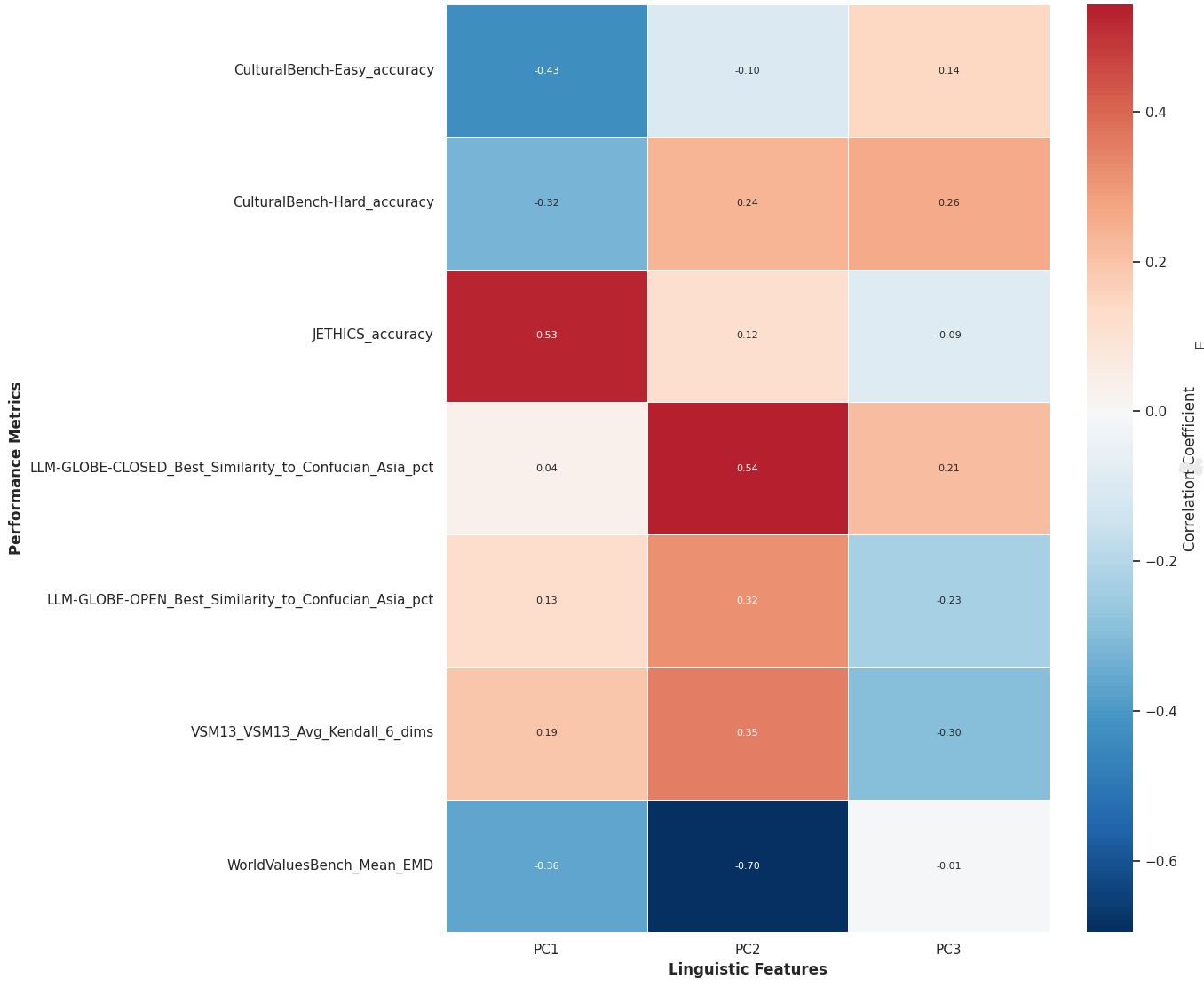}
        \caption*{(c) JP–DeepSeek}
    \end{minipage}

    \caption{Correlation between dataset PCA components (PC1--PC3) and downstream cultural performance for Japanese across models.}
    \label{fig:japanese_corr_heatmaps}
\end{figure*}

\begin{figure*}[htbp]
    \centering

    \begin{minipage}[t]{0.32\textwidth}
        \centering
        \includegraphics[width=\linewidth]{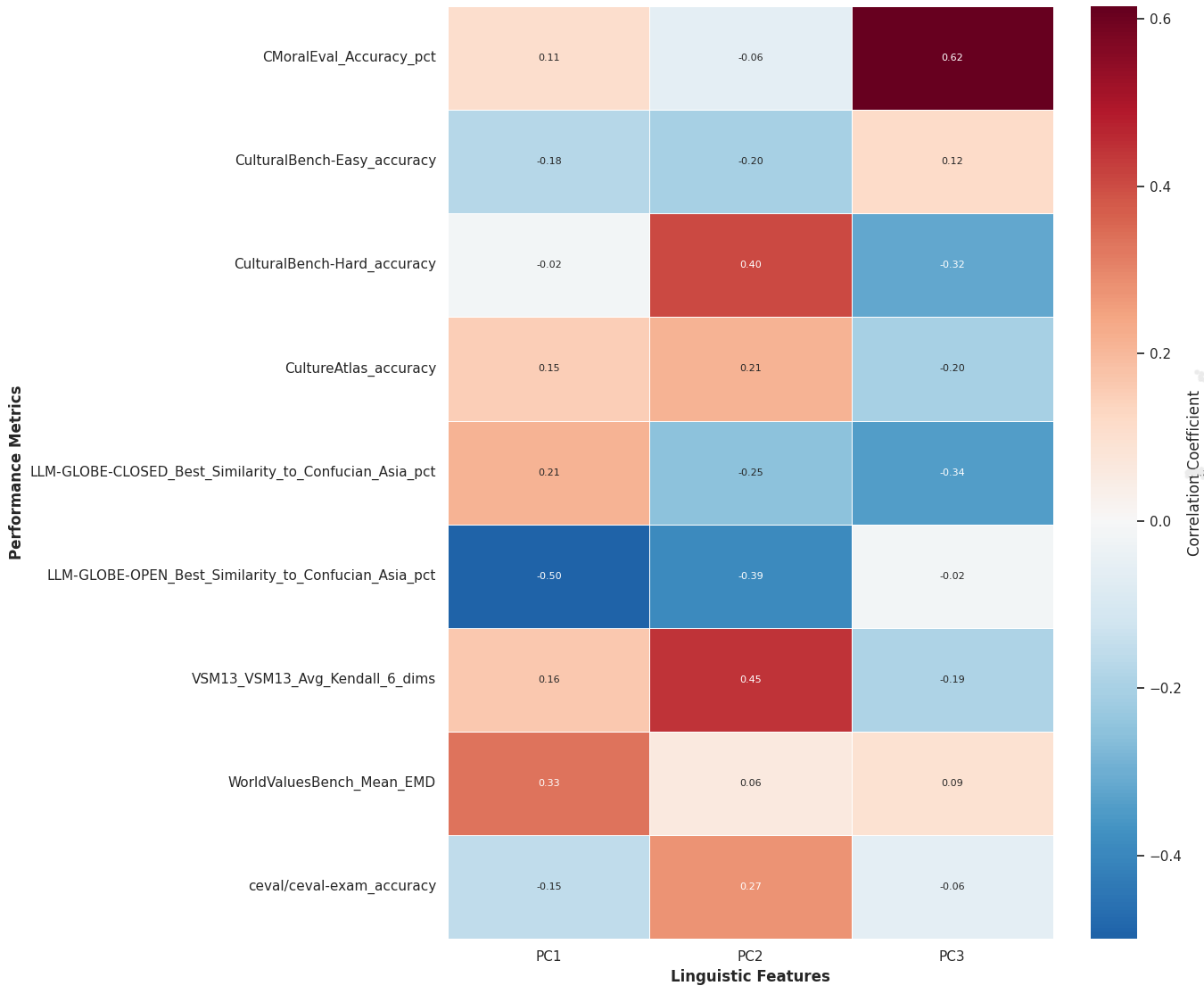}
        \caption*{(a) ZH–LLaMA}
    \end{minipage}\hfill
    \begin{minipage}[t]{0.32\textwidth}
        \centering
        \includegraphics[width=\linewidth]{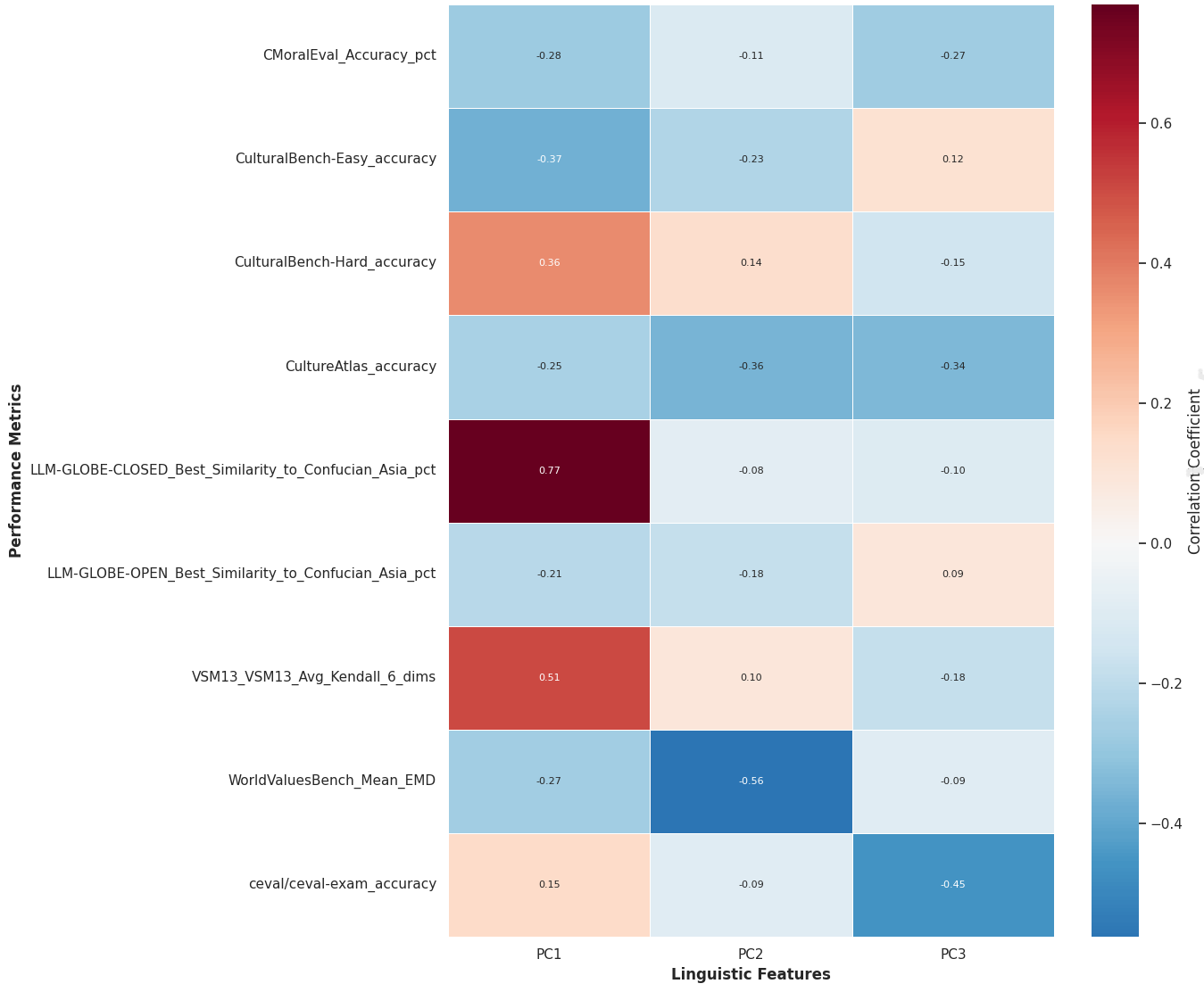}
        \caption*{(b) ZH–Mistral}
    \end{minipage}\hfill
    \begin{minipage}[t]{0.32\textwidth}
        \centering
        \includegraphics[width=\linewidth]{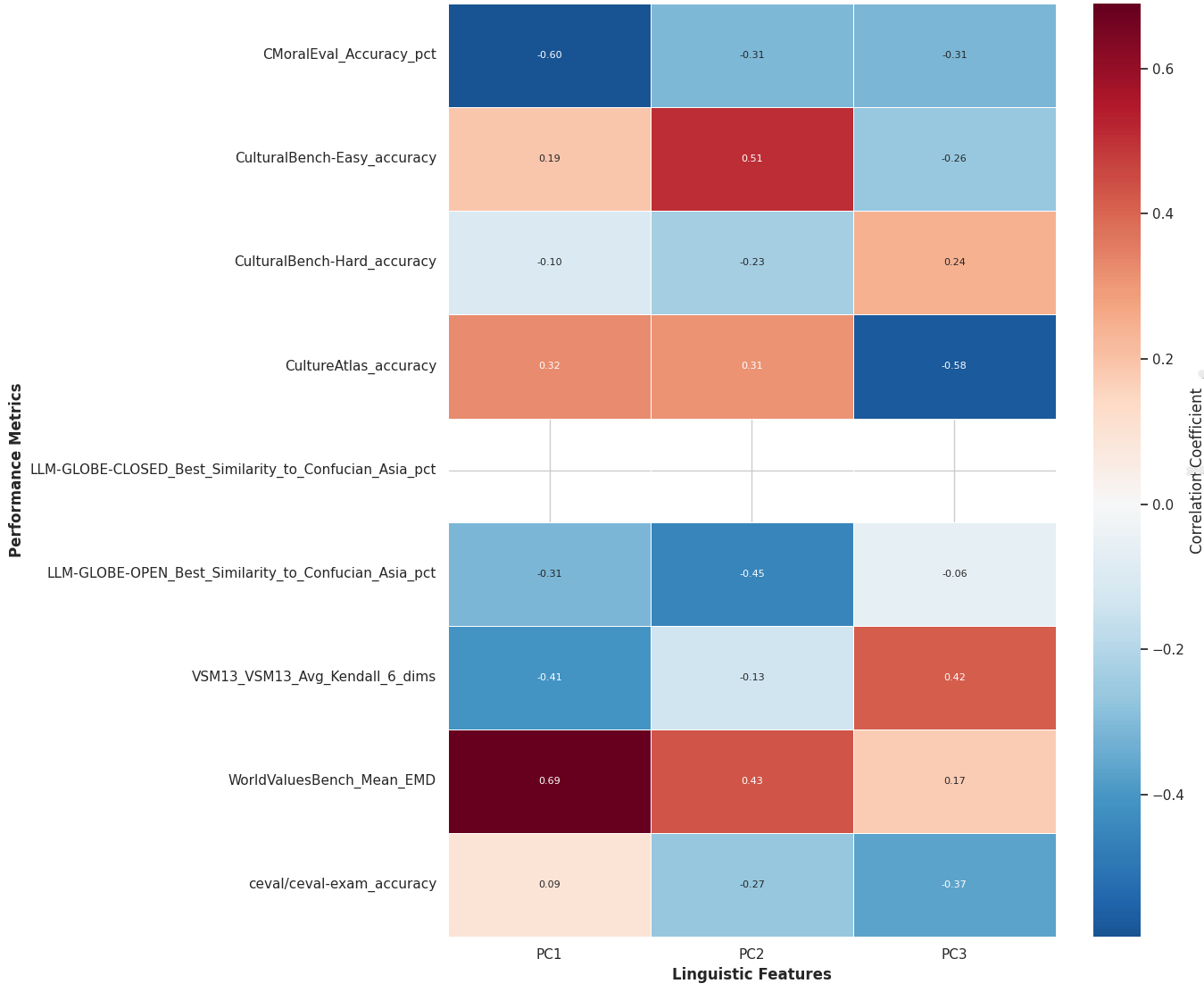}
        \caption*{(c) ZH–DeepSeek}
    \end{minipage}

    \caption{Correlation between dataset PCA components (PC1--PC3) and downstream cultural performance for Chinese across models.}
    \label{fig:chinese_corr_heatmaps}
\end{figure*}

This appendix presents the full correlation heatmaps between dataset PCA components (PC1–PC3) and downstream cultural benchmark performance, reported separately for each language and model. Each cell in Figs.~\ref{fig:arabic_corr_heatmaps}, \ref{fig:japanese_corr_heatmaps}, and \ref{fig:chinese_corr_heatmaps} shows the correlation coefficient between a PCA component score (columns) and a benchmark score (rows).

\section{Detailed Controlled-Subset Results}
\label{appendix:subset_results}

The main paper reports the aggregate directional outcomes of the controlled
subset interventions. This section provides the underlying model- and
source-dataset-level results. For each language and principal component,
the five source datasets with the largest signed dataset-level component
scores are selected. Within each source dataset, size-matched
\textsc{High-PC}, \textsc{Low-PC}, and \textsc{Random} subsets are
constructed and used to fine-tune each model independently under the
training configuration specified in the main paper.

We report two complementary analyses. The directional figures classify each
source-dataset intervention as an \emph{Improvement}, \emph{Tie}, or
\emph{Decline}, according to whether benchmark-level wins outnumber losses,
losses outnumber wins, or the counts are equal. The normalized-margin figures
retain the scale-adjusted magnitude of the performance difference relative
to the corresponding Random subset.

\subsection{Directional Outcomes}
\label{appendix:directional_results}

\begin{figure*}[p]
    \centering
    \includegraphics[
        width=\textwidth,
        height=0.88\textheight,
        keepaspectratio
    ]{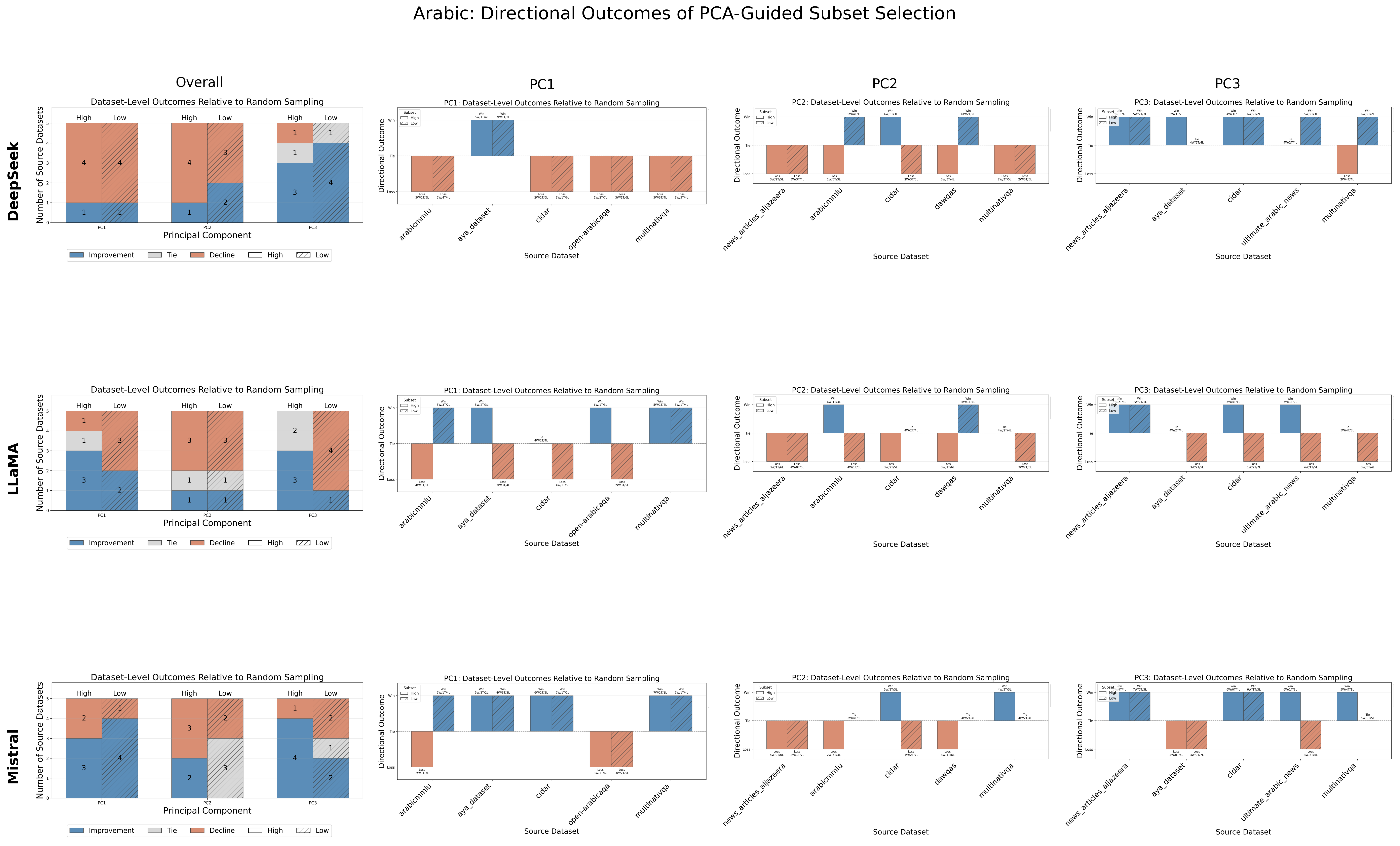}
    \caption{Detailed directional outcomes of PCA-guided subset selection
    for Arabic. Rows correspond to DeepSeek, LLaMA, and Mistral. The
    Overall column reports dataset-level Improvement/Tie/Decline
    classifications across the five selected source datasets. The remaining
    columns show source-dataset outcomes for PC1--PC3. Solid bars denote
    High-PC subsets and hatched bars denote Low-PC subsets.}
    \label{fig:appendix_arabic_directional}
\end{figure*}

\begin{figure*}[p]
    \centering
    \includegraphics[
        width=\textwidth,
        height=0.88\textheight,
        keepaspectratio
    ]{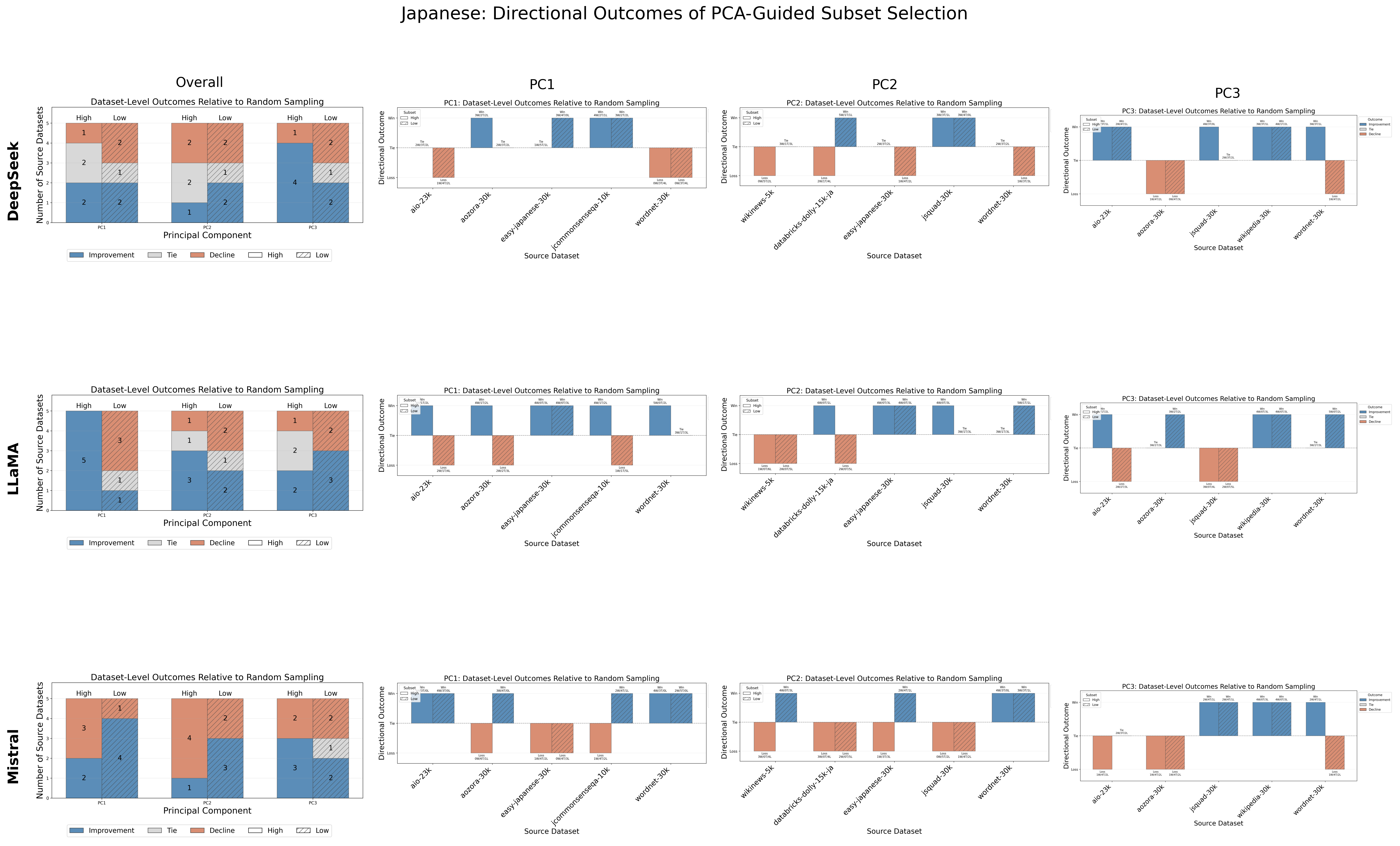}
    \caption{Detailed directional outcomes of PCA-guided subset selection
    for Japanese. Rows correspond to DeepSeek, LLaMA, and Mistral. The
    Overall column reports dataset-level Improvement/Tie/Decline
    classifications across the five selected source datasets. The remaining
    columns show source-dataset outcomes for PC1--PC3. Solid bars denote
    High-PC subsets and hatched bars denote Low-PC subsets.}
    \label{fig:appendix_japanese_directional}
\end{figure*}

\begin{figure*}[p]
    \centering
    \includegraphics[
        width=\textwidth,
        height=0.88\textheight,
        keepaspectratio
    ]{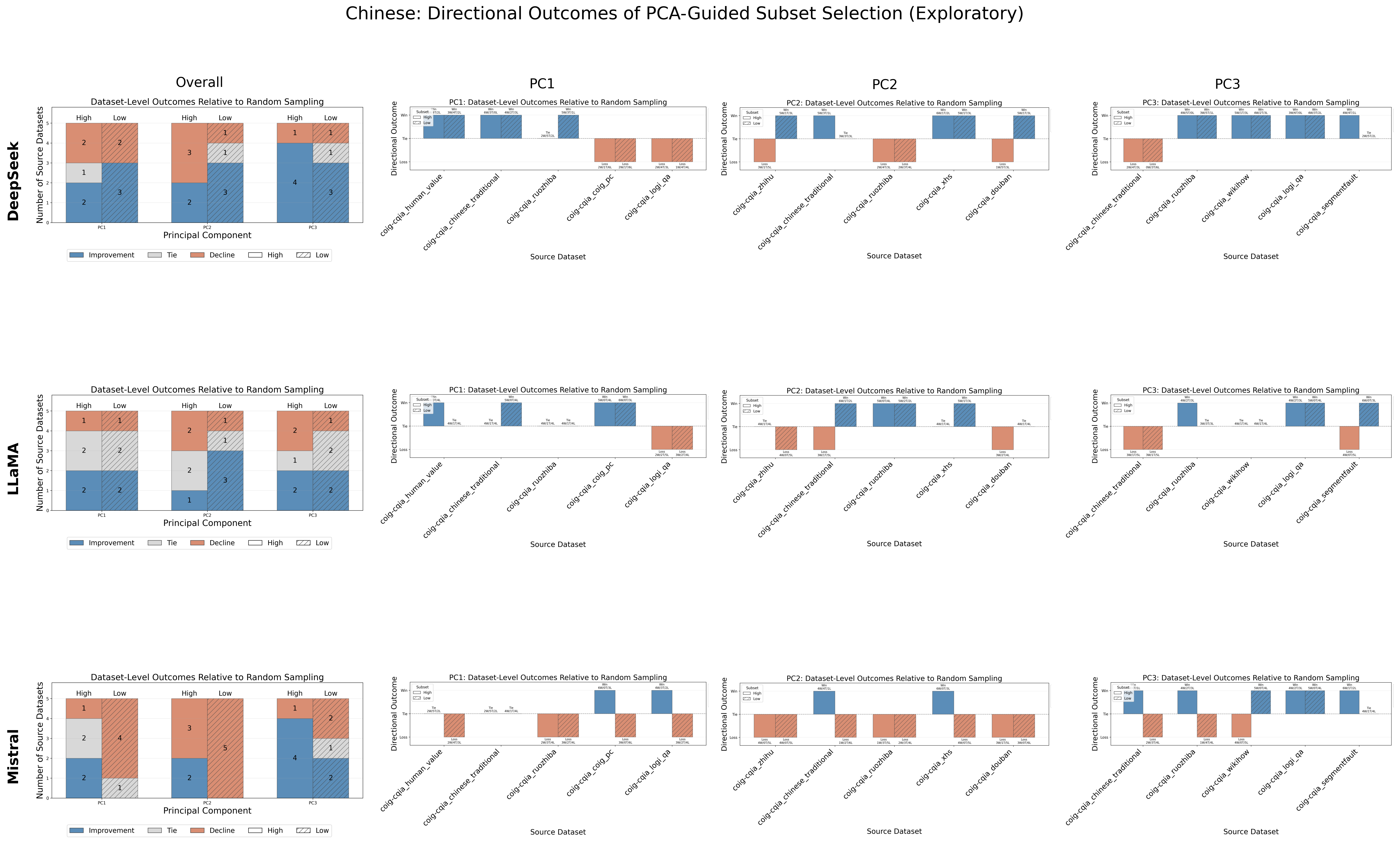}
    \caption{Detailed directional outcomes of PCA-guided subset selection
    for Chinese. Rows correspond to DeepSeek, LLaMA, and Mistral. The
    Overall column reports dataset-level Improvement/Tie/Decline
    classifications across the five selected source datasets. The remaining
    columns show source-dataset outcomes for PC1--PC3. Solid bars denote
    High-PC subsets and hatched bars denote Low-PC subsets. Chinese results
    are treated as exploratory because the source datasets and resulting
    subsets are smaller.}
    \label{fig:appendix_chinese_directional}
\end{figure*}

Figures~\ref{fig:appendix_arabic_directional}--
\ref{fig:appendix_chinese_directional} provide the complete directional
outcomes. In each figure, rows correspond to model families. The
\emph{Overall} column reports the number of the five source-dataset
interventions classified as Improvement, Tie, or Decline. The remaining
columns report the underlying source-dataset outcomes for PC1--PC3. Solid
bars denote High-PC subsets and hatched bars denote Low-PC subsets. The
source-dataset triplets are benchmark-level win/tie/loss counts, whereas the
Overall bars report dataset-level Improvement/Tie/Decline classifications.
These labels describe observed directional recurrence and do not imply
statistical significance.

\clearpage
\subsection{Magnitude of the Intervention Effects}
\label{appendix:magnitude_results}

The directional analysis measures how recurrently a PCA-guided subset
outperforms its size-matched Random baseline, but it does not retain the
magnitude of the performance differences. We therefore additionally report
the mean normalized margin for each source-dataset intervention. For
benchmark $b$, model $m$, language $\ell$, source dataset $d$, and guided
condition $s\in\{\mathrm{High},\mathrm{Low}\}$, the benchmark-level normalized
margin is

\begin{equation}
\Delta^{\mathrm{norm}}_{\ell,m,d,s,b}
=
\frac{
y_{\ell,m,d,s,b}
-
y_{\ell,m,d,\mathrm{Random},b}
}{
R_{\ell,m,b}
},
\end{equation}

where $y_{\ell,m,d,s,b}$ is the direction-aligned benchmark score and
$R_{\ell,m,b}$ is the observed non-zero range of benchmark $b$ within the
same language--model setting. The mean normalized margin is

\begin{equation}
\overline{\Delta}^{\mathrm{norm}}_{\ell,m,d,s}
=
\frac{1}{|\mathcal{B}_{\ell,m,d,s}|}
\sum_{b\in\mathcal{B}_{\ell,m,d,s}}
\Delta^{\mathrm{norm}}_{\ell,m,d,s,b},
\end{equation}

where $\mathcal{B}_{\ell,m,d,s}$ contains the available benchmarks with a
non-zero observed range. Positive values indicate better average performance
than Random sampling, negative values indicate worse average performance,
and zero represents the Random baseline.

For each model and principal component, the figures report separate
horizontal diverging bars for the High-PC and Low-PC strategies. Bars
extending to the right indicate positive mean normalized margins, while bars
extending to the left indicate negative margins. A common horizontal scale is
used across panels to support comparison between components and subset
directions. Unlike the directional analysis, a comparatively large effect on
one benchmark can offset several smaller effects in the opposite direction.
The directional and magnitude analyses should therefore be interpreted
jointly: agreement indicates that an effect is both recurrent and positive in
average magnitude, whereas disagreement distinguishes the breadth of an
effect from its numerical size.

\subsubsection{Arabic}

\begin{figure*}[p]
    \centering
    \includegraphics[
        width=\textwidth,
        height=0.84\textheight,
        keepaspectratio
    ]{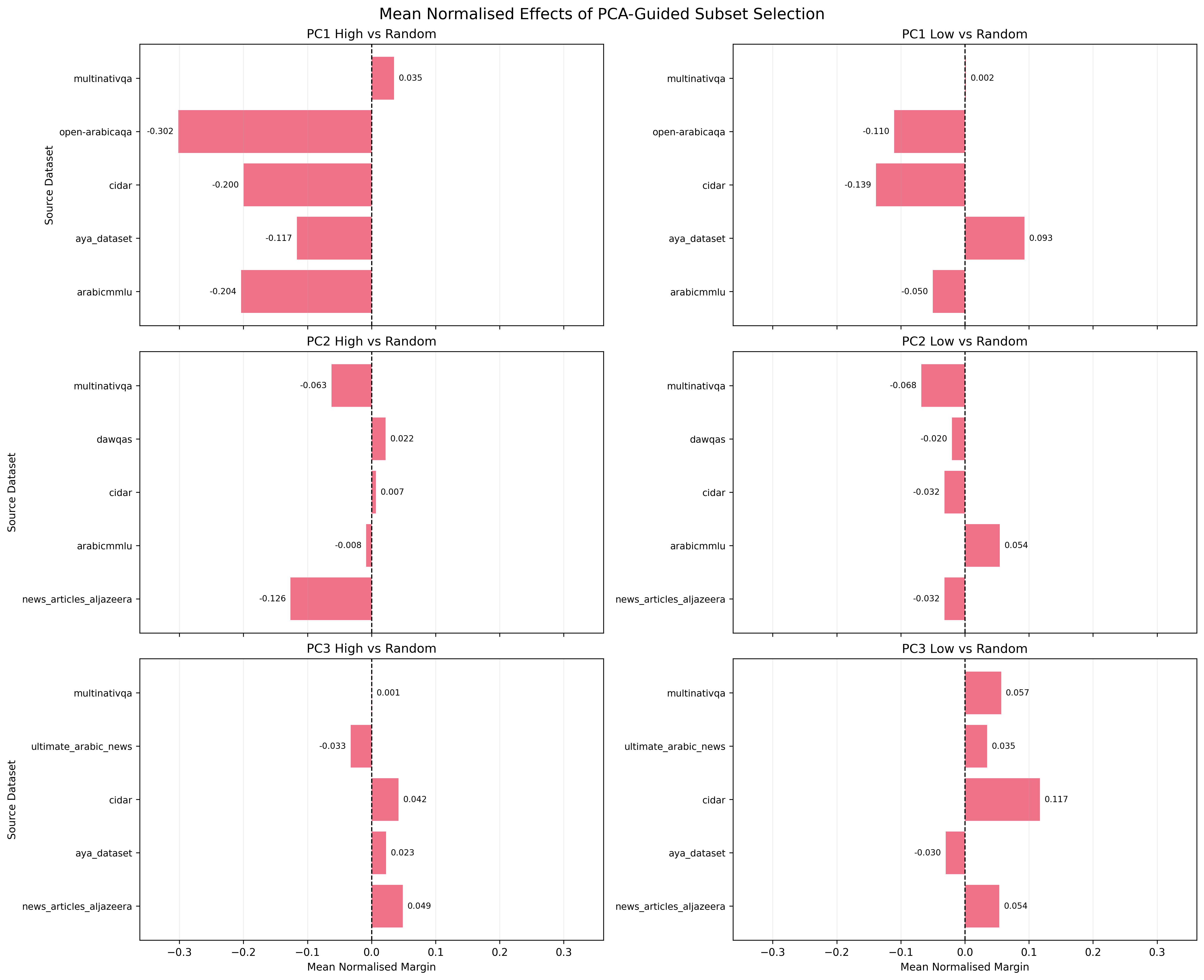}
    \caption{
        Mean normalized Arabic subset effects for DeepSeek relative to
        size-matched Random sampling. Positive values indicate average
        improvement, while negative values indicate average decline.
    }
    \label{fig:appendix_arabic_magnitude_deepseek}
\end{figure*}

\begin{figure*}[p]
    \centering
    \includegraphics[
        width=\textwidth,
        height=0.84\textheight,
        keepaspectratio
    ]{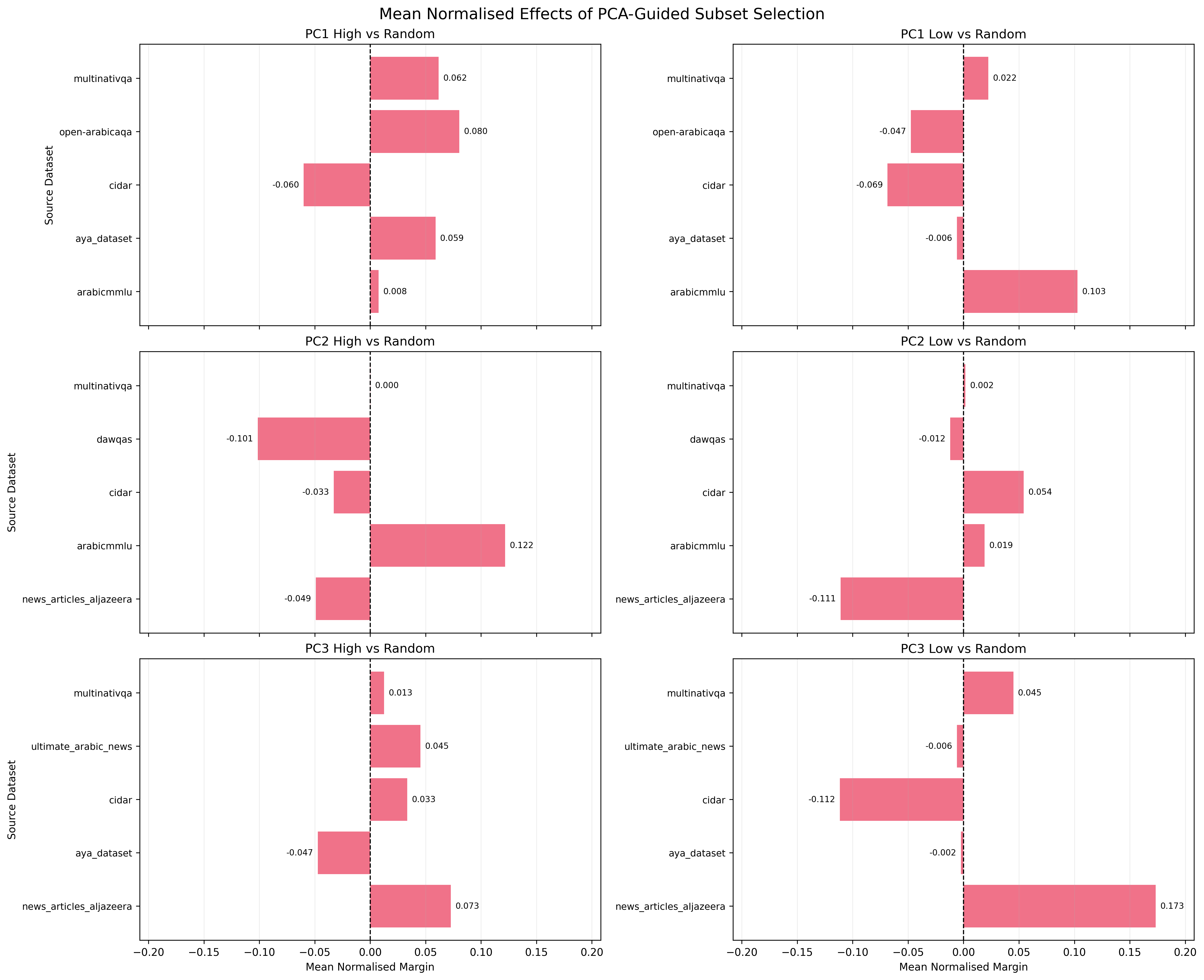}
    \caption{
        Mean normalized Arabic subset effects for LLaMA relative to
        size-matched Random sampling.
    }
    \label{fig:appendix_arabic_magnitude_llama}
\end{figure*}

\begin{figure*}[p]
    \centering
    \includegraphics[
        width=\textwidth,
        height=0.84\textheight,
        keepaspectratio
    ]{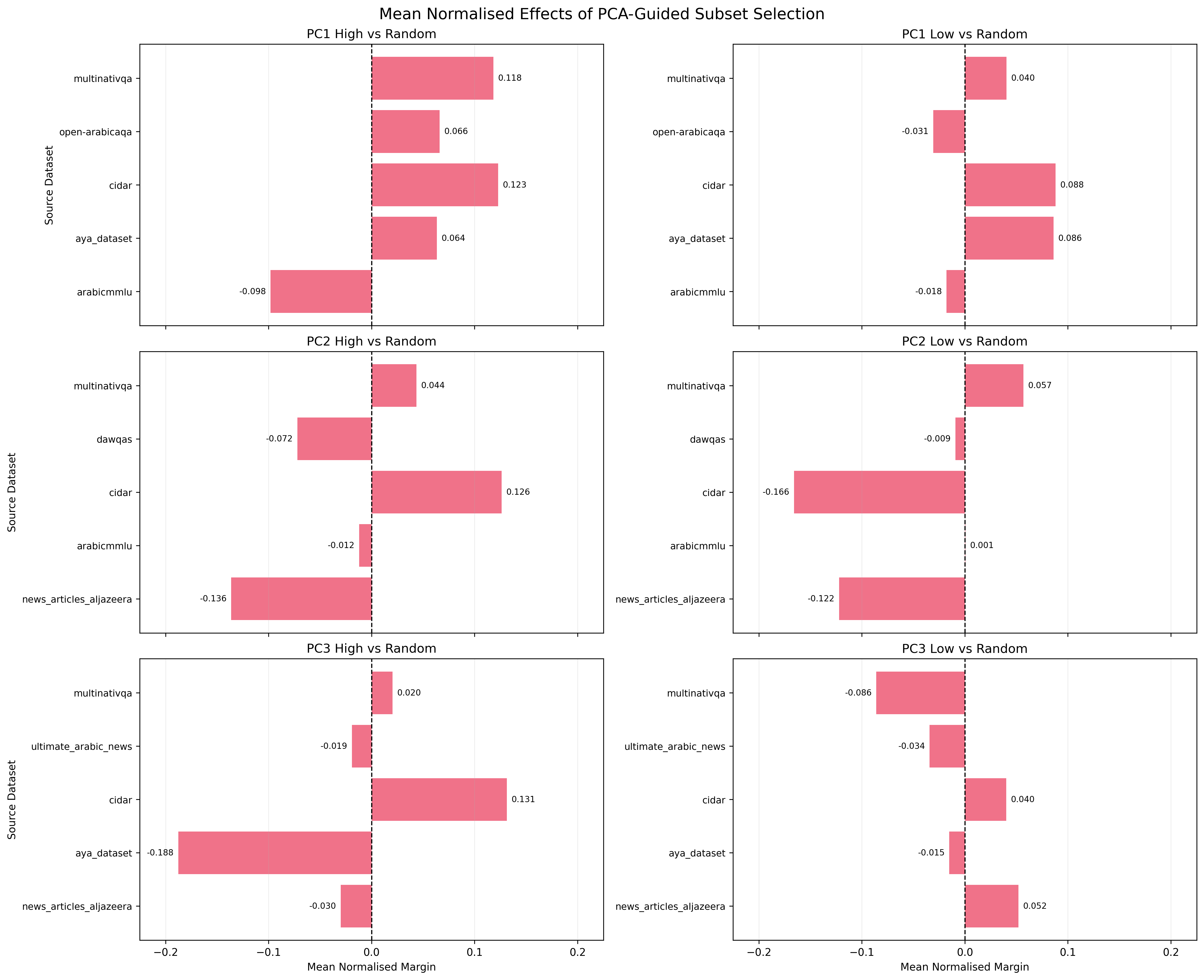}
    \caption{
        Mean normalized Arabic subset effects for Mistral relative to
        size-matched Random sampling.
    }
    \label{fig:appendix_arabic_magnitude_mistral}
\end{figure*}

Figures~\ref{fig:appendix_arabic_magnitude_deepseek}--
\ref{fig:appendix_arabic_magnitude_mistral} report the Arabic
source-dataset margins.

PC1 produces some of the largest Arabic effects, but its direction differs
substantially across model families. DeepSeek shows large negative margins
under High-PC1, whereas LLaMA and Mistral exhibit more favorable High-PC1
effects for several source datasets. PC1 therefore represents a high-impact
but architecture-dependent selection dimension.

PC2 produces the weakest overall magnitude pattern. Across the three models,
High-PC2 and Low-PC2 yield mixed and generally small or negative effects,
consistent with the directional analysis identifying PC2 as the least
reliable Arabic selection dimension.

PC3 provides the strongest Arabic magnitude evidence overall. DeepSeek shows
positive effects under both High-PC3 and Low-PC3, with Low-PC3 producing the
larger margins. LLaMA shows a broadly positive High-PC3 pattern, while
Mistral also produces several positive High-PC3 margins, although these are
partly offset by the negative effect for Aya. Together with the directional
results, these findings identify PC3 as the most consistently favorable
Arabic selection dimension, while PC1 is higher-impact but less stable.

\clearpage
\subsubsection{Japanese}

\begin{figure*}[p]
    \centering
    \includegraphics[
        width=\textwidth,
        height=0.84\textheight,
        keepaspectratio
    ]{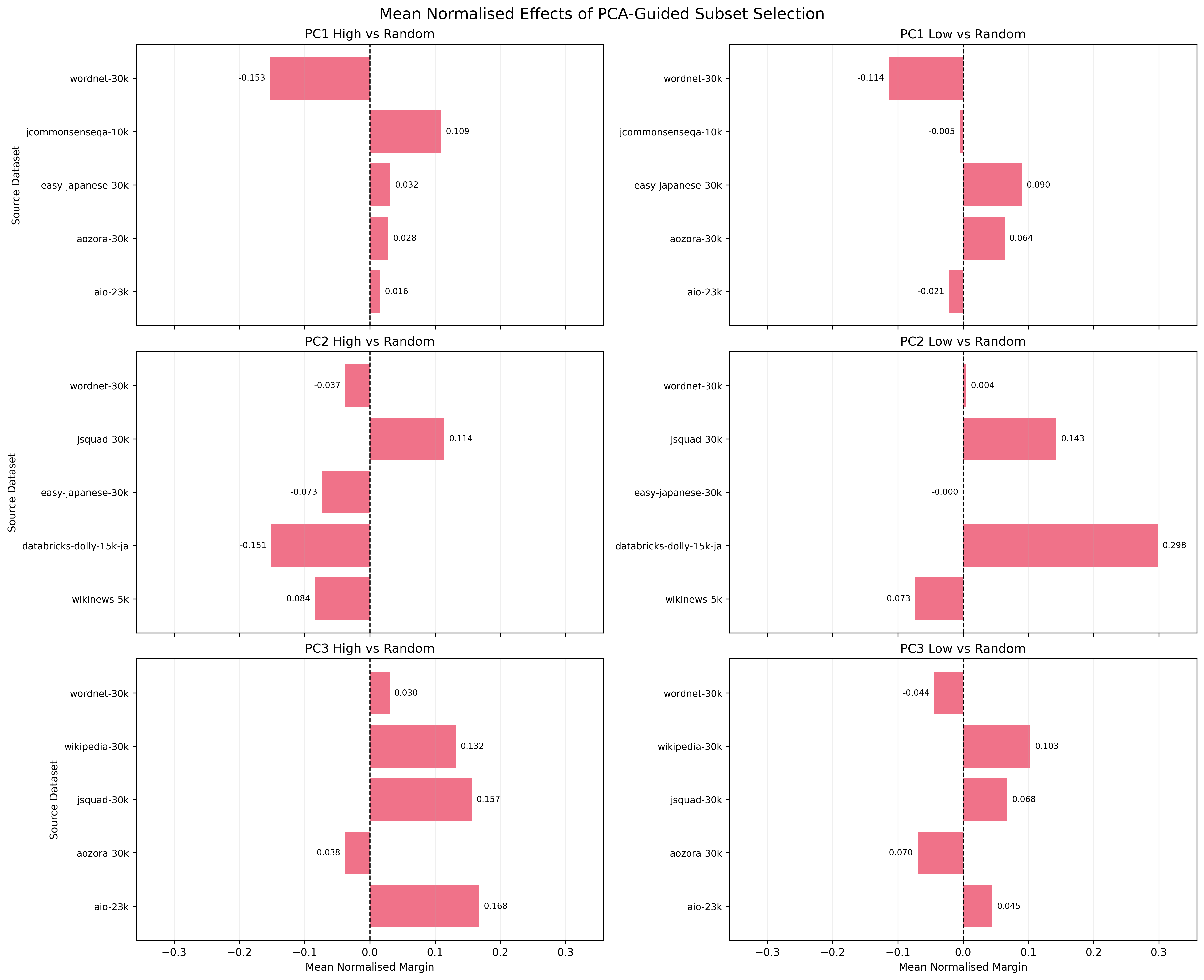}
    \caption{
        Mean normalized Japanese subset effects for DeepSeek relative to
        size-matched Random sampling.
    }
    \label{fig:appendix_japanese_magnitude_deepseek}
\end{figure*}

\begin{figure*}[p]
    \centering
    \includegraphics[
        width=\textwidth,
        height=0.84\textheight,
        keepaspectratio
    ]{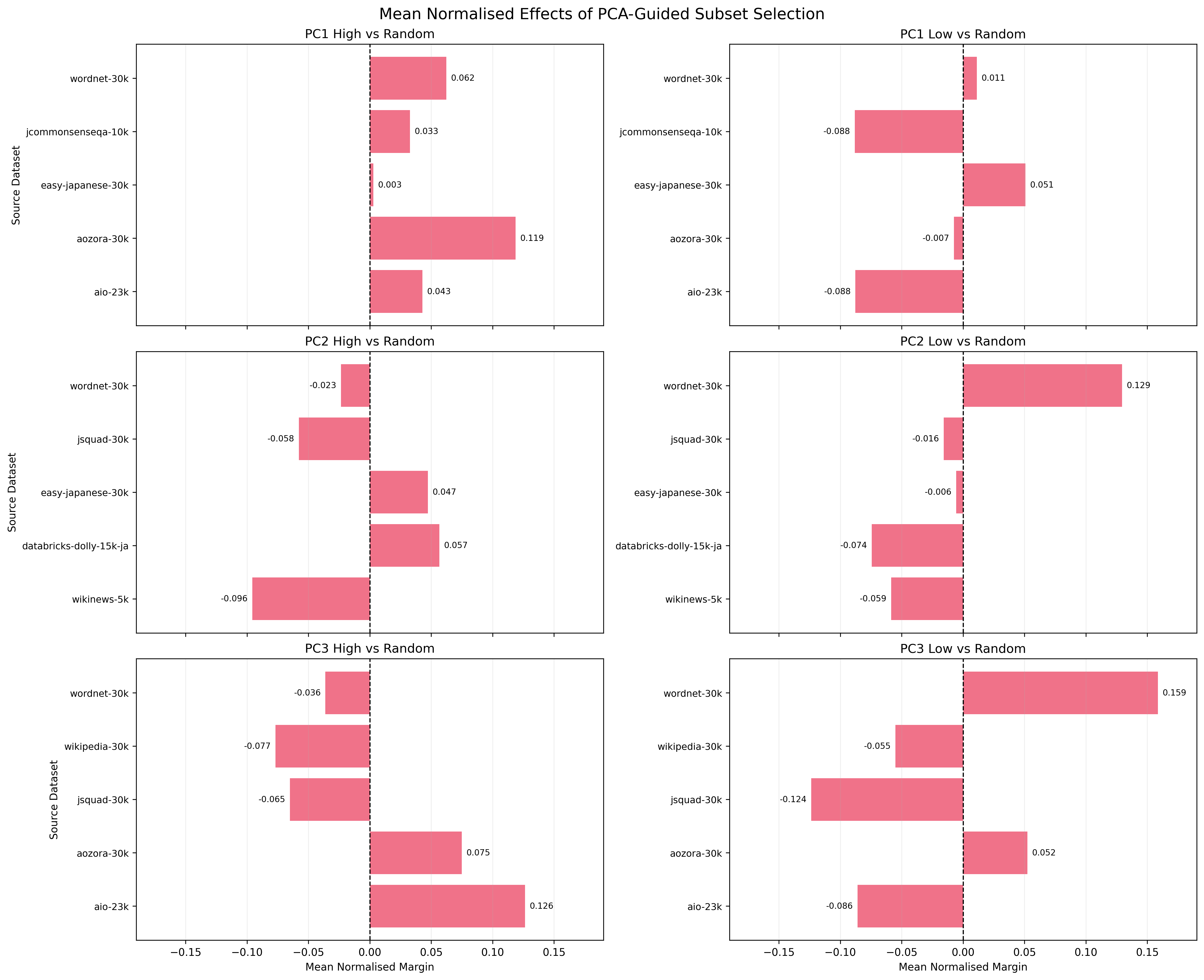}
    \caption{
        Mean normalized Japanese subset effects for LLaMA relative to
        size-matched Random sampling.
    }
    \label{fig:appendix_japanese_magnitude_llama}
\end{figure*}

\begin{figure*}[p]
    \centering
    \includegraphics[
        width=\textwidth,
        height=0.84\textheight,
        keepaspectratio
    ]{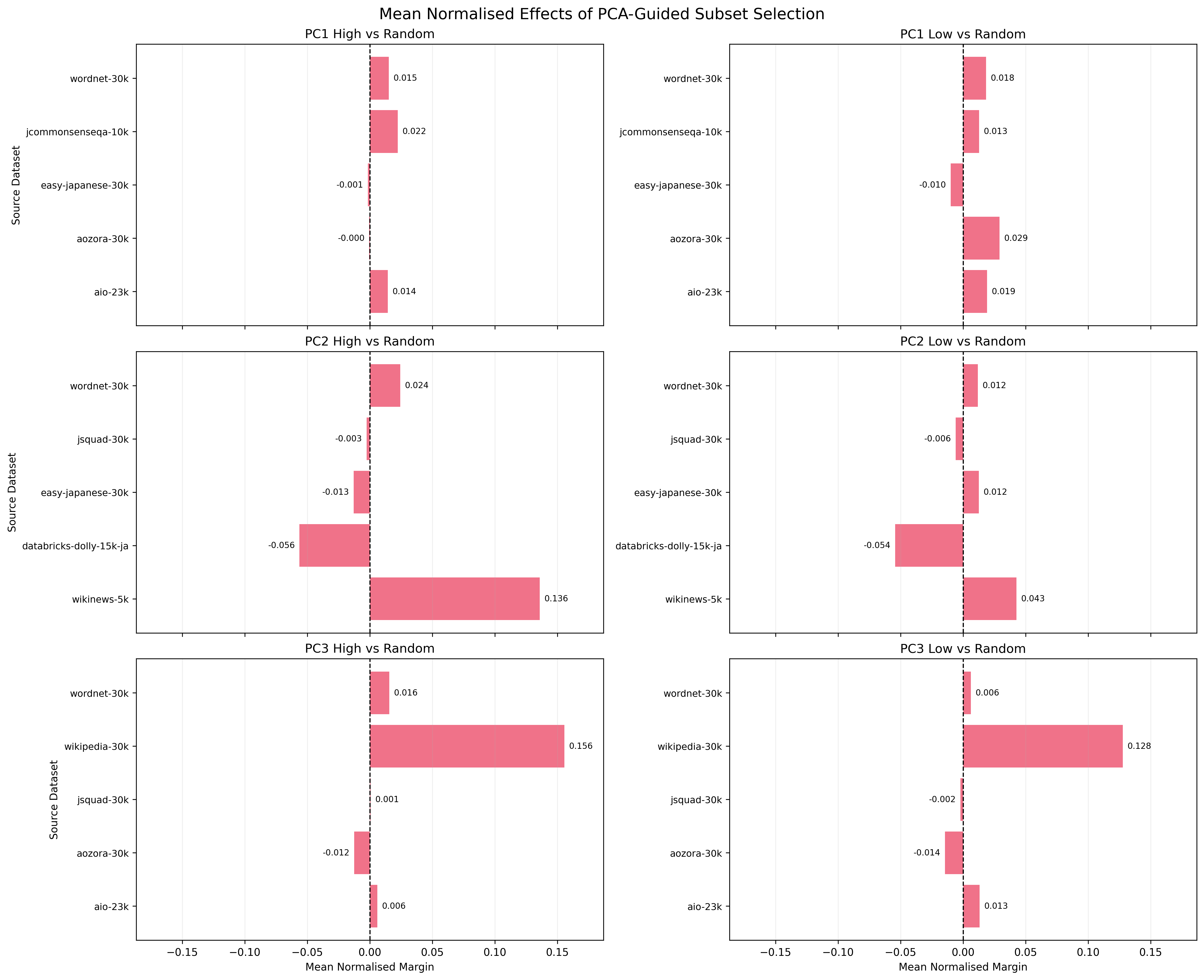}
    \caption{
        Mean normalized Japanese subset effects for Mistral relative to
        size-matched Random sampling.
    }
    \label{fig:appendix_japanese_magnitude_mistral}
\end{figure*}

Figures~\ref{fig:appendix_japanese_magnitude_deepseek}--
\ref{fig:appendix_japanese_magnitude_mistral} report the Japanese
source-dataset margins.

PC1 provides the clearest positive magnitude evidence for LLaMA. High-PC1
produces positive margins across all five source datasets, matching the
directional analysis in which all five interventions were classified as
improvements. DeepSeek and Mistral show more variable PC1 effects, including
large negative margins for WordNet. PC1 therefore provides the strongest
Japanese selection signal for LLaMA, but its effect does not transfer
uniformly across architectures.

PC2 is strongly source-dataset dependent. DeepSeek and Mistral obtain large
positive Low-PC2 effects for some datasets, particularly Databricks Dolly and
JSQuAD, but these gains coexist with negative margins elsewhere. PC2 can
therefore produce substantial model--dataset-specific gains without providing
a stable cross-dataset selection rule.

PC3 produces several comparatively large positive margins despite being
predominantly tie-dominated in the directional analysis. DeepSeek shows a
broadly positive High-PC3 pattern, while LLaMA and Mistral exhibit larger
effects concentrated in particular source datasets. Thus, PC3 can produce
substantial gains, but these gains are not sufficiently recurrent to define a
general selection rule.

Overall, the Japanese magnitude results support High-PC1 as the clearest
positive selection pattern, particularly for LLaMA, while showing that PC3
can still yield large source-dataset-specific improvements.

\clearpage
\subsubsection{Chinese}

The Chinese intervention is interpreted as exploratory because its source
datasets and resulting size-matched subsets are smaller than those used for
Arabic and Japanese.

\begin{figure*}[p]
    \centering
    \includegraphics[
        width=\textwidth,
        height=0.84\textheight,
        keepaspectratio
    ]{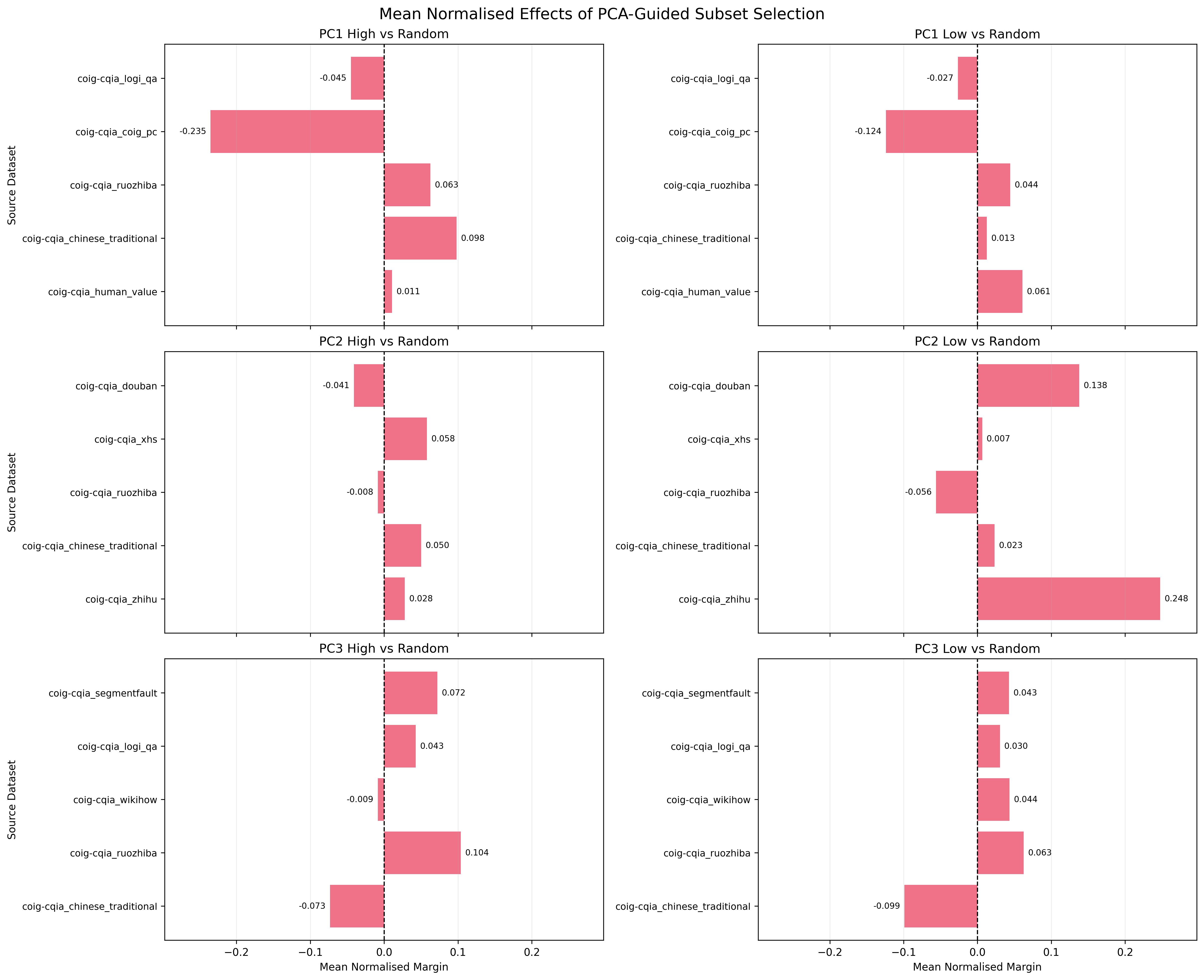}
    \caption{
        Mean normalized Chinese subset effects for DeepSeek relative to
        size-matched Random sampling.
    }
    \label{fig:appendix_chinese_magnitude_deepseek}
\end{figure*}

\begin{figure*}[p]
    \centering
    \includegraphics[
        width=\textwidth,
        height=0.84\textheight,
        keepaspectratio
    ]{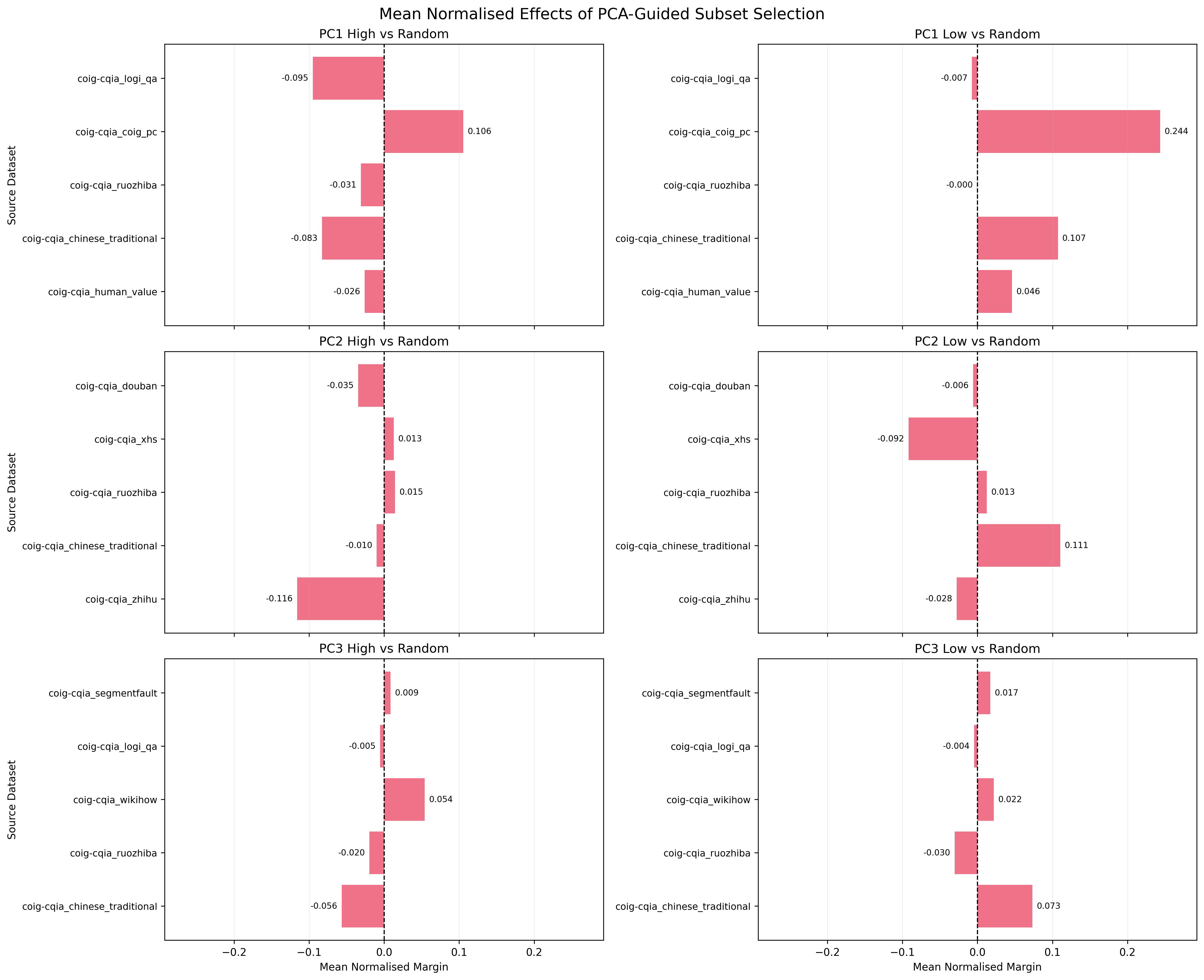}
    \caption{
        Mean normalized Chinese subset effects for LLaMA relative to
        size-matched Random sampling.
    }
    \label{fig:appendix_chinese_magnitude_llama}
\end{figure*}

\begin{figure*}[p]
    \centering
    \includegraphics[
        width=\textwidth,
        height=0.84\textheight,
        keepaspectratio
    ]{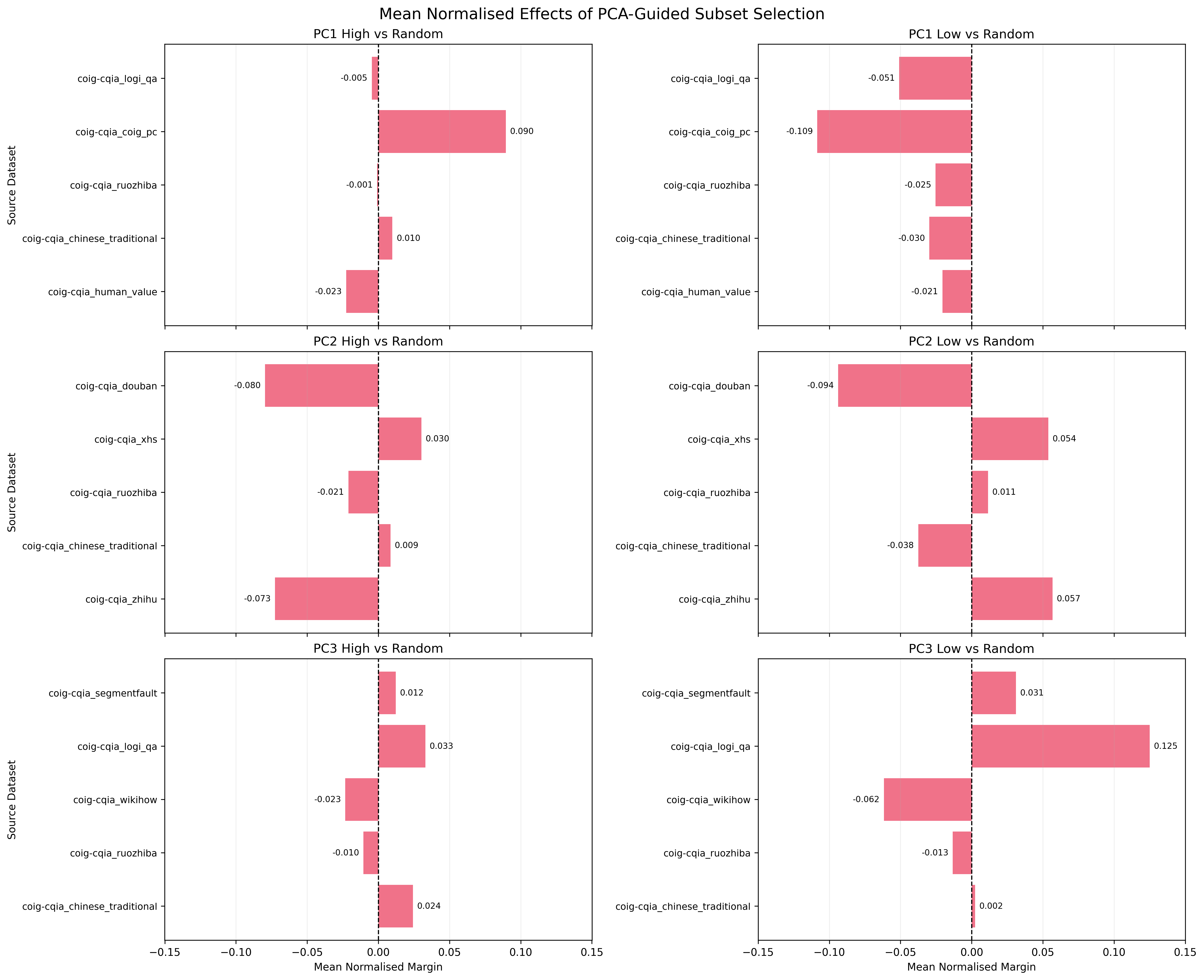}
    \caption{
        Mean normalized Chinese subset effects for Mistral relative to
        size-matched Random sampling.
    }
    \label{fig:appendix_chinese_magnitude_mistral}
\end{figure*}

Figures~\ref{fig:appendix_chinese_magnitude_deepseek}--
\ref{fig:appendix_chinese_magnitude_mistral} report the Chinese
source-dataset margins.

PC1 remains strongly architecture dependent. DeepSeek is mixed under both
subset directions, while LLaMA shows its clearest positive pattern under
Low-PC1. In contrast, all five Low-PC1 margins for Mistral are negative.
PC1 therefore does not provide a transferable cross-model selection rule,
although Low-PC1 is favorable for LLaMA.

PC2 also varies across architectures. DeepSeek responds most favorably to
Low-PC2, which produces positive margins for four of the five source
datasets. LLaMA is mixed under both directions, while Mistral produces
predominantly negative High-PC2 effects and a mixed Low-PC2 pattern. The
clearest PC2 magnitude signal is therefore specific to Low-PC2 for DeepSeek.

PC3 produces mixed but generally more favorable effects. DeepSeek shows its
strongest PC3 pattern under Low-PC3, with four of the five source datasets
producing positive margins. LLaMA remains more variable, while Mistral shows
several positive effects under both directions, including its largest gain
under Low-PC3. Although the directional analysis characterizes PC3 as
predominantly neutral, the magnitude results indicate that it can produce
meaningful improvements for particular model--dataset combinations.

Taken together, the Chinese magnitude results reveal clearer numerical
patterns than the directional counts alone. PC3 and Low-PC2 provide the
strongest signals for DeepSeek, Low-PC1 is clearest for LLaMA, and PC3 is
most favorable for Mistral. These results remain exploratory because of the
smaller Chinese datasets and subset sizes.

\clearpage
\subsection{Summary of Detailed Intervention Results}

The detailed magnitude results refine the aggregate directional findings
reported in the main paper. For Arabic, PC3 shows the clearest agreement
between recurrent improvements and positive effect magnitude, although the
preferred subset direction differs across models. PC1 produces some larger
effects but is substantially less stable, while PC2 provides the weakest
overall evidence.

For Japanese, High-PC1 provides the strongest recurrent and positive
within-dataset selection signal, particularly for LLaMA. PC3 is largely
tie-dominated in the directional analysis but can nevertheless produce
substantial gains for particular source datasets, illustrating the difference
between recurrence and effect magnitude.

The Chinese results are weaker, more architecture dependent, and interpreted
as exploratory. Low-PC2 and PC3 provide the clearest signals for DeepSeek,
Low-PC1 is most favorable for LLaMA, and PC3 produces the strongest positive
effects for Mistral.

Overall, PCA-guided selection is most useful as a language-, model-, and
source-dataset-aware diagnostic procedure. Neither the identity of the
principal component nor the High-PC or Low-PC direction should be treated as
a universal indicator of example quality.

\section{The Use of Artificial Intelligence}

In the development of this paper, we employed artificial intelligence
(AI) tools to enhance writing quality and grammatical accuracy. All
research decisions, analyses, interpretations, and final manuscript content
were reviewed and approved by the authors.

\end{document}